\documentclass[preprint]{elsarticle}

\usepackage{lineno,hyperref}
\usepackage{amsmath}
\usepackage{times}
\usepackage{epsfig}
\usepackage{graphicx}
\usepackage{amssymb}
\newcommand{\etal}{\textit{et al.}}
\modulolinenumbers[5]

\journal{ISPRS Journal of Photogrammetry and Remote Sensing}

\begin{document}

\begin{frontmatter}

\title{Advanced Underwater Image Restoration in Complex Illumination Conditions}

\author[add1]{Yifan Song\corref{correspondingauthor}} 
\cortext[correspondingauthor]{Corresponding author}
\ead{ysong@geomar.de}

\author[add1,add2]{Mengkun She} 
\author[add1,add2]{Kevin K{\"o}ser}

\address[add1]{Oceanic Machine Vision, GEOMAR Helmholtz Centre for Ocean Research Kiel, \\Wischhofstr. 1-3, 24148 Kiel, Germany}
\address[add2]{Marine Data Science, Christian-Albrechts-University of Kie, \\Neufeldtstr. 6, 24118 Kiel, Germany}

%\author{Yifan Song, Mengkun She, Kevin K{\"o}ser}%\fnref{myfootnote}}
%\address{Oceanic Machine Vision, GEOMAR Helmholtz Centre for Ocean Research Kiel, \\Wischhofstr. 1-3, 24148 Kiel}

\begin{abstract}
	
Underwater image restoration has been a challenging problem for decades since the advent of underwater photography. 
Most solutions focus on shallow water scenarios, where the scene is uniformly illuminated by the sunlight.
However, the vast majority of uncharted underwater terrain is located beyond 200 meters depth where natural light is scarce and artificial illumination is needed. 
In such cases, light sources co-moving with the camera, dynamically change the scene appearance, which make shallow water restoration methods inadequate.
In particular for multi-light source systems (composed of dozens of LEDs nowadays), calibrating each light is time-consuming, error-prone and tedious, and we observe that only the integrated illumination within the viewing volume of the camera is critical, rather than the individual light sources. 
The key idea of this paper is therefore to exploit the appearance changes of objects or the seafloor, when traversing the viewing frustum of the camera. Through new constraints assuming Lambertian surfaces, corresponding image pixels constrain the light field in front of the camera, and for each voxel a signal factor and a backscatter value are stored in a volumetric grid that can be used for very efficient image restoration of camera-light platforms, which facilitates consistently texturing large 3D models and maps that would otherwise be dominated by lighting and medium artifacts. To validate the effectiveness of our approach, we conducted extensive experiments on simulated and real-world datasets. The results of these experiments demonstrate the robustness of our approach in restoring the true albedo of objects, while mitigating the influence of lighting and medium effects. Furthermore, we demonstrate our approach can be readily extended to other scenarios, including in-air imaging with artificial illumination or other similar cases. 

\end{abstract}


\end{frontmatter}

%\linenumbers

\section{Introduction}
\label{sec:introduction}

Water covers about 70\% of the Earth‘s surface, but only very limited portion of the seafloor has been explored and charted. 
With the increasing interest in ocean research and exploration, visual mapping of the seafloor using camera vision systems is becoming more popular. %due to their advantages for human interpretation. 
However, the majority of the seafloor is situated below the Mesopelagic zone where nature light cannot penetrates, requiring additional artificial illumination during the imaging.
Unlike images in the shallow water, the appearance of deep water images is significantly influenced by the lighting configurations. 
Unfortunately, current underwater image processing solutions mostly focus on shallow water cases with homogeneous illumination and are not applicable to images under complex illumination conditions. 
With the developments of underwater robotics, we are able to explore the deepest regions of the ocean, and a more general restoration solution for different types of underwater images is increasingly demanded. 

\begin{figure}
\centering
\includegraphics[width=0.9\linewidth]{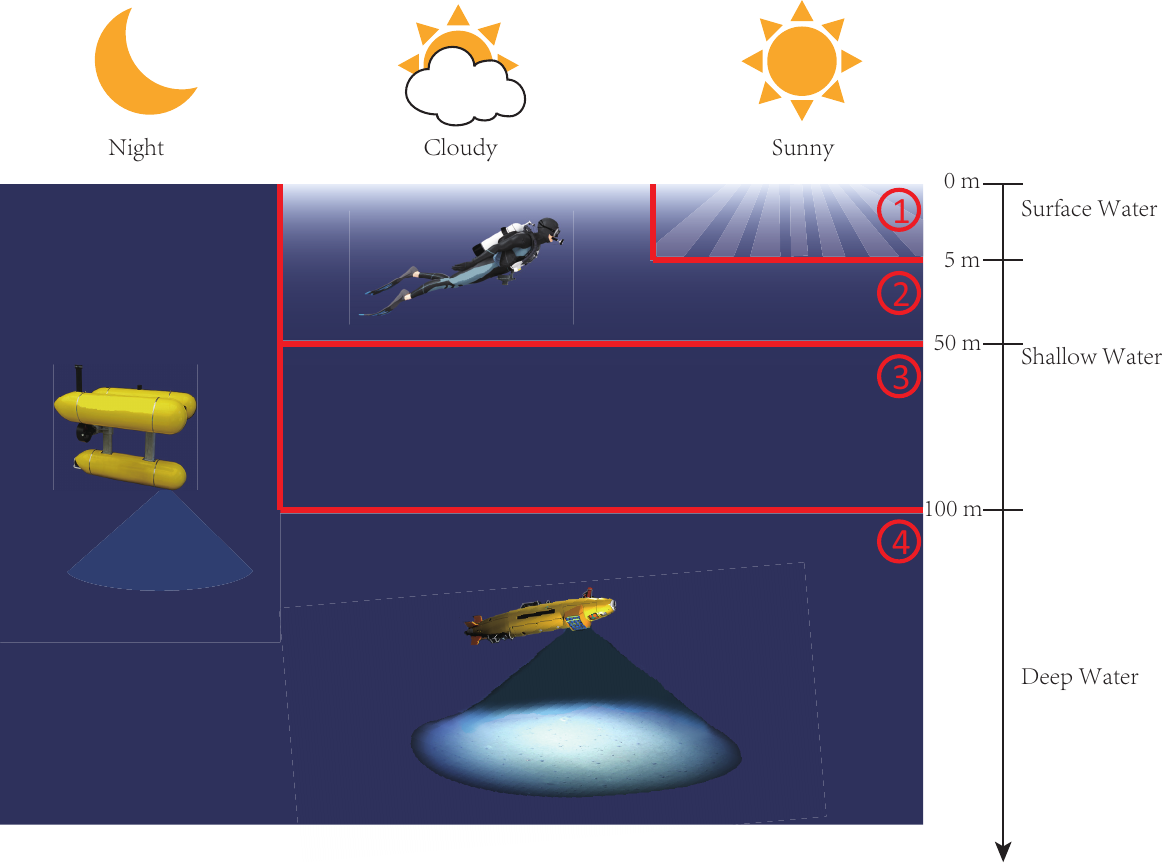}
\caption{Different image formation models under different illumination conditions.}
\label{fig_diff_illumination}
\end{figure}

In underwater imaging, this paper classifies the physical-based underwater image formation models into four categories according to their illumination conditions (see Fig. \ref{fig_diff_illumination}). 

\begin{figure}
\centering
\includegraphics[width=0.45\linewidth]{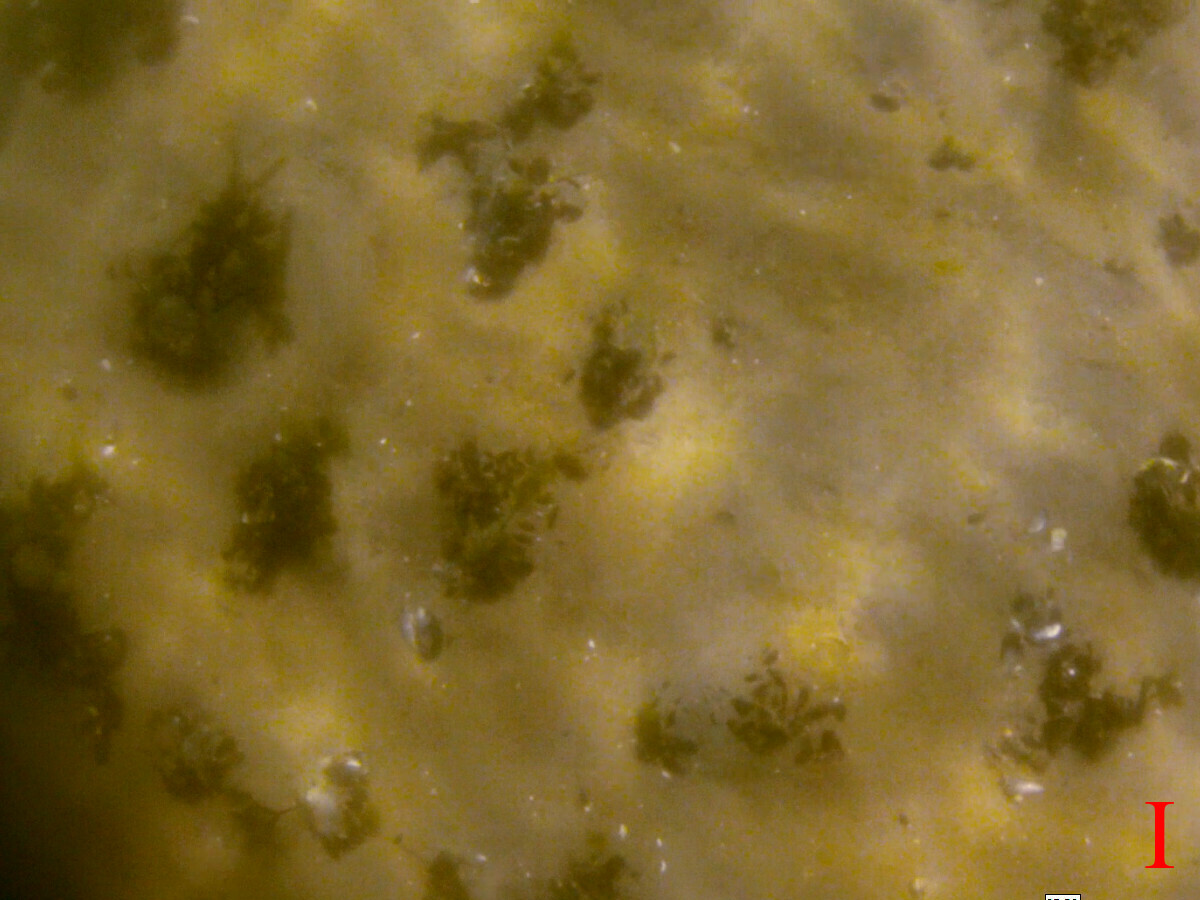}
\includegraphics[width=0.45\linewidth]{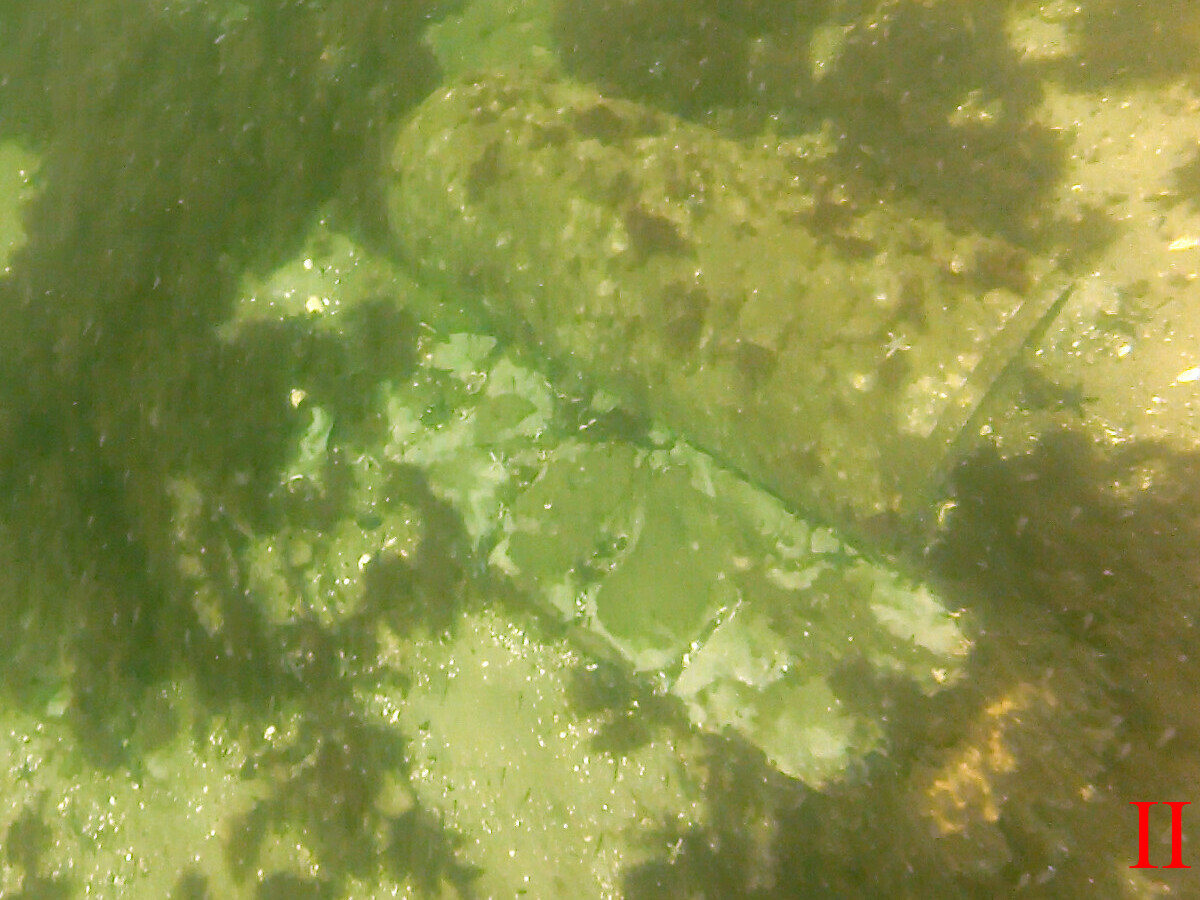} 
\\ [1mm]
\includegraphics[width=0.45\linewidth]{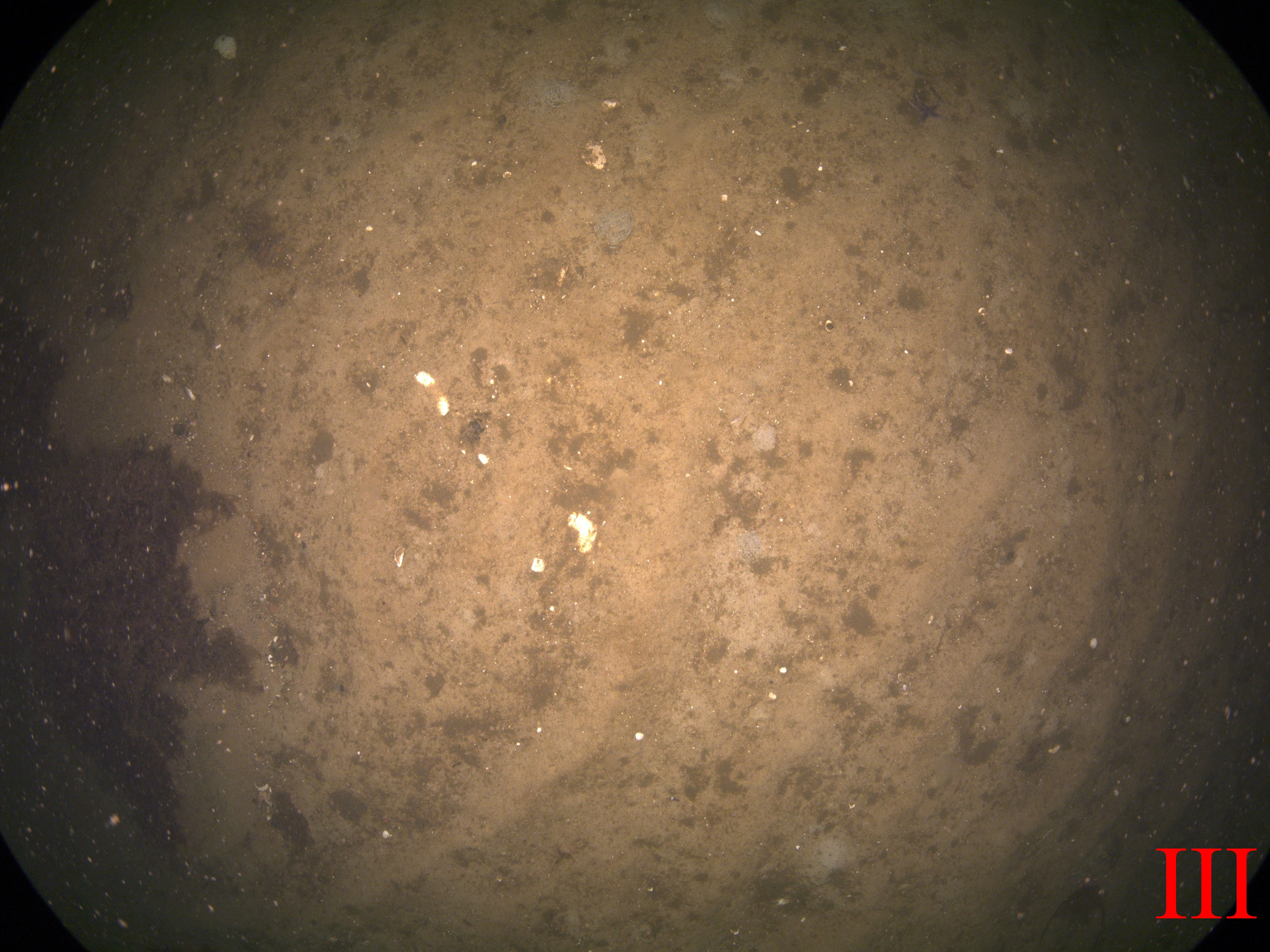} 
\includegraphics[width=0.45\linewidth]{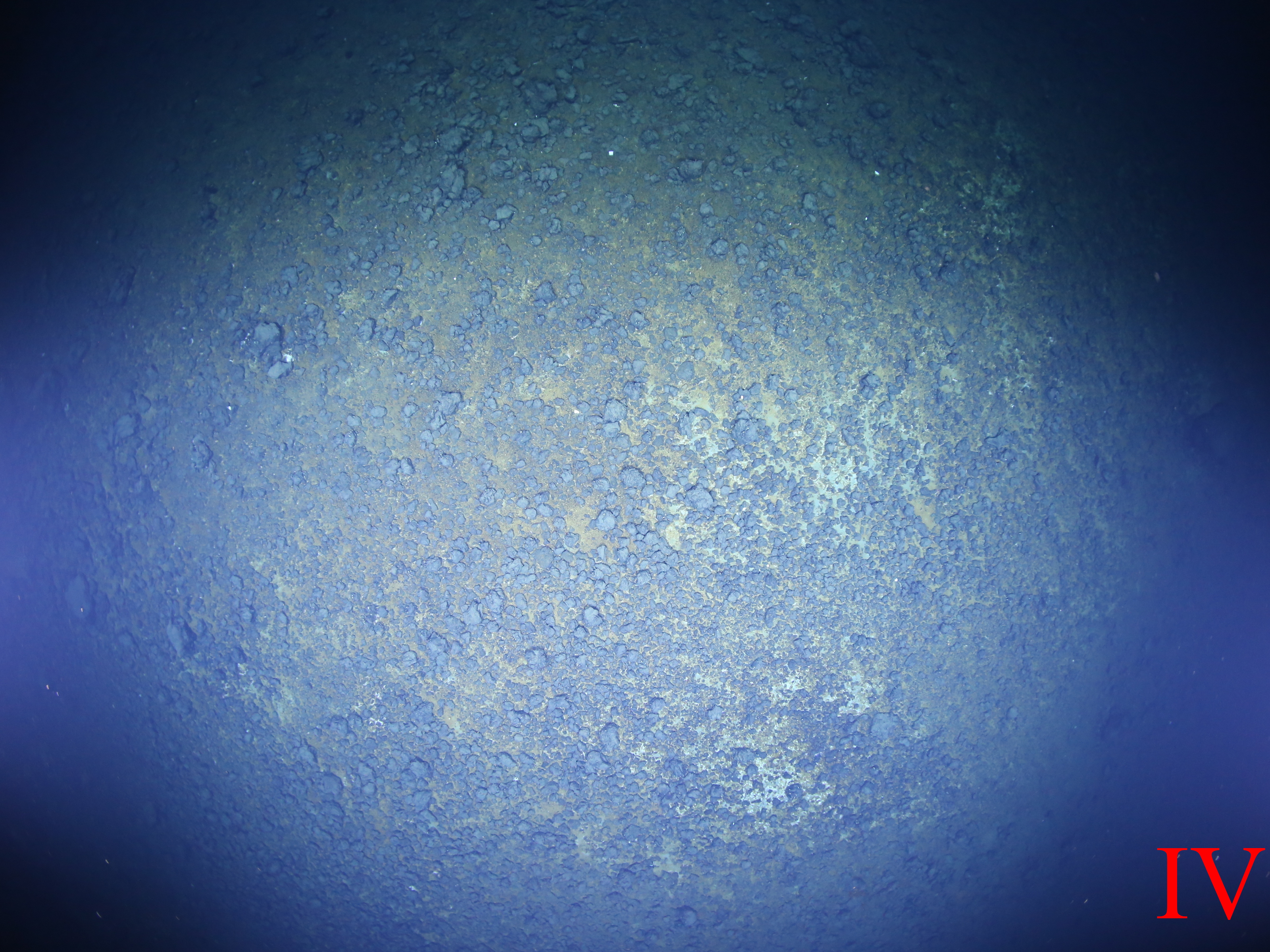}

\caption{Examples of underwater seafloor images captured under different illumination conditions, each corresponding to a different image formation model. \textbf{I}: In surface water where the strong sunlight creates a dynamic caustic pattern. \textbf{II}: In shallow water where the illumination is relatively homogeneous due to the abundant sunlight. \textbf{III}: in the twilight zone where the sunlight is severely attenuated and additional artificial light is used to illuminate the scene. \textbf{IV}: In complete darkness in the depth ocean and is illuminated solely by artificial light sources.}
\label{fig_diff_illumination_images}
\end{figure}

\begin{itemize}
\item \textbf{Type I}: \emph{Surface water model}. 
This model describes the image formation in the water surface region where the scene is completely illuminated by sunlight. 
Its basic image formation model is similar to Type II, but strong sunlight is refracted dynamically at waves of the water surface, producing additional caustic patterns in the scene. 
The caustic patterns are constantly changing due to the water surface and it is challenging to predict the caustic pattern in the water based on physical models, as it requires information such as water surface normal, water depth, geometry of the scene and the relative position of the sun. 

\item \textbf{Type II}: \emph{Shallow water model}. 
This model is by far the most popular model which been widely applied in underwater image processing methods. It descends from atmospheric dehazing which originally been used to recover the depth cues from images affected by haze or fog. In this model the underwater image formation is composed by direct attenuated light and ambient light(backscatter). 
The sunlight first travels from the water surface to the seafloor, and then be reflected to the camera. 
The attenuation of the sunlight in the first path, known as veiling light, requires knowledge of the water depth, but the attenuated light in the same region of water is relatively homogeneous, allowing for it to be approximated as the background color. 
The attenuation of object intensity is only considered in the second path, resulting in the corresponding image formation model becomes a weighted linear combination of object intensity and background color (backscatter). 
%atmospheric scattering model, some parallel viewing image also use: the ocean-\-atmosphere radiative transfer (OSOA) model \cite{chami2001radiative}
%Type IIa, downward looking, constant background color, fog model.
%Type IIb, near horizontal view, gradient change background color along depth, fog + OSOA model.

\item \textbf{Type III}: \emph{Mixed model}. 
This type of model combines characteristics of both Type II and IV models. 
While the nature sunlight is not enough to illuminate the scene, the ambient illumination is not completely dark, and thus additional artificial illumination is required to supplement the illumination.

\item \textbf{Type IV}: \emph{Deep water model}. 
When the region is devoid of sunlight, the scene is illuminated solely by artificial light sources co-moving with the camera. In this image formation model, the signal is still a sum of direct and backscattered light (forward scattering effect is often approximated as the extra smooth over the signal). However, the attenuation of light in water now needs to consider the path from the artificial light source to the object and then to the camera. Meanwhile, the artificial light sources have different spectrum to sunlight, which must also be taken into account.
The total backscatter in the scene is no longer represented by a single, uniform background color. Instead, it is an integral of water body scattering along each viewing ray, which depends on the configuration of the artificial illumination water properties such as the Volume Scattering Function (VSF). 
The most popular model is the Jaffe-McGlamery model. 
\end{itemize}
Example images for each type are illustrated in Fig. \ref{fig_diff_illumination_images}.

%-------------------------------------------------------------------------
\section{Related work and main contributions}

Underwater image restoration for seafloor mapping involves addressing several issues such as recovering attenuated color, removing backscatter, homogenizing lighting pattern (if artificial illumination is present) and maintaining color consistency of the same object across images. 
In this context, we provide a brief overview of related work, while a more comprehensive review is available in our previous publication \cite{song2022optical}. 

The pioneer work began  in the domain of atmospheric scattering, where attempts were made to recover depth information from images captured in fog or haze.
\cite{cozman1997depth} brought the atmospheric scattering model from physics to computer vision and extracted depth cues from the scattering effects present in the images. This model describes the atmospheric scattering image formation as a weighted linear combination of object intensity $I_0$ and sky intensity $S$: 

\begin{equation}
\label{FogModel}
I = e^{-\eta d} \cdot I_0 + (1-e^{-\eta d}) \cdot S.
\end{equation}

The exponential term indicates the decreases of the signal in the medium according to the environmental attenuation coefficient $\eta$ and its traveling path $d$, while also approximates the increase of the backscatter (background light). This model has been adapted in many physical model-based in-air image dehazing approaches \cite{narasimhan2003contrast, tan2008visibility, he2010single, zhu2015fast, berman2016non}. 
Similarly, these concepts have been applied in the underwater domain \cite{sedlazeck20093d,drews2013transmission,berman2020underwater}. Upon examining the details of these methods, we noticed that most physical model based underwater image restoration methods can be generalized as solving the estimation of transmission term $T$ and backscatter term $B$ in:

\begin{equation}
\label{eqSimpleModel}
I = T \cdot I_0 + B.
\end{equation}

Restoring color from single image is an ill pose problem. The estimation of transmission and backscatter terms can be solved by introducing extra prior knowledge constraints or through multiple correspondence observations. 
Prior constraints aim to discover distance-related changes in the single image to recover the transmission and backscatter terms for each pixel. Popular priors include the Dark Channel Prior \cite{he2010single} and its derivatives \cite{carlevaris2010initial,drews2013transmission,peng2018generalization}, the Haze-Lines Prior \cite{berman2016non,berman2020underwater} and the Blurriness Prior \cite{peng2015single,peng2017underwater}. However, the quality of the results from prior knowledge-based methods depends on the image content itself and cannot guarantee consistent output over large image sequences for mapping purpose. Moreover, they are not able to deal with strong artificial lighting patterns. 

\begin{figure}
\centering
\includegraphics[width=0.8\linewidth]{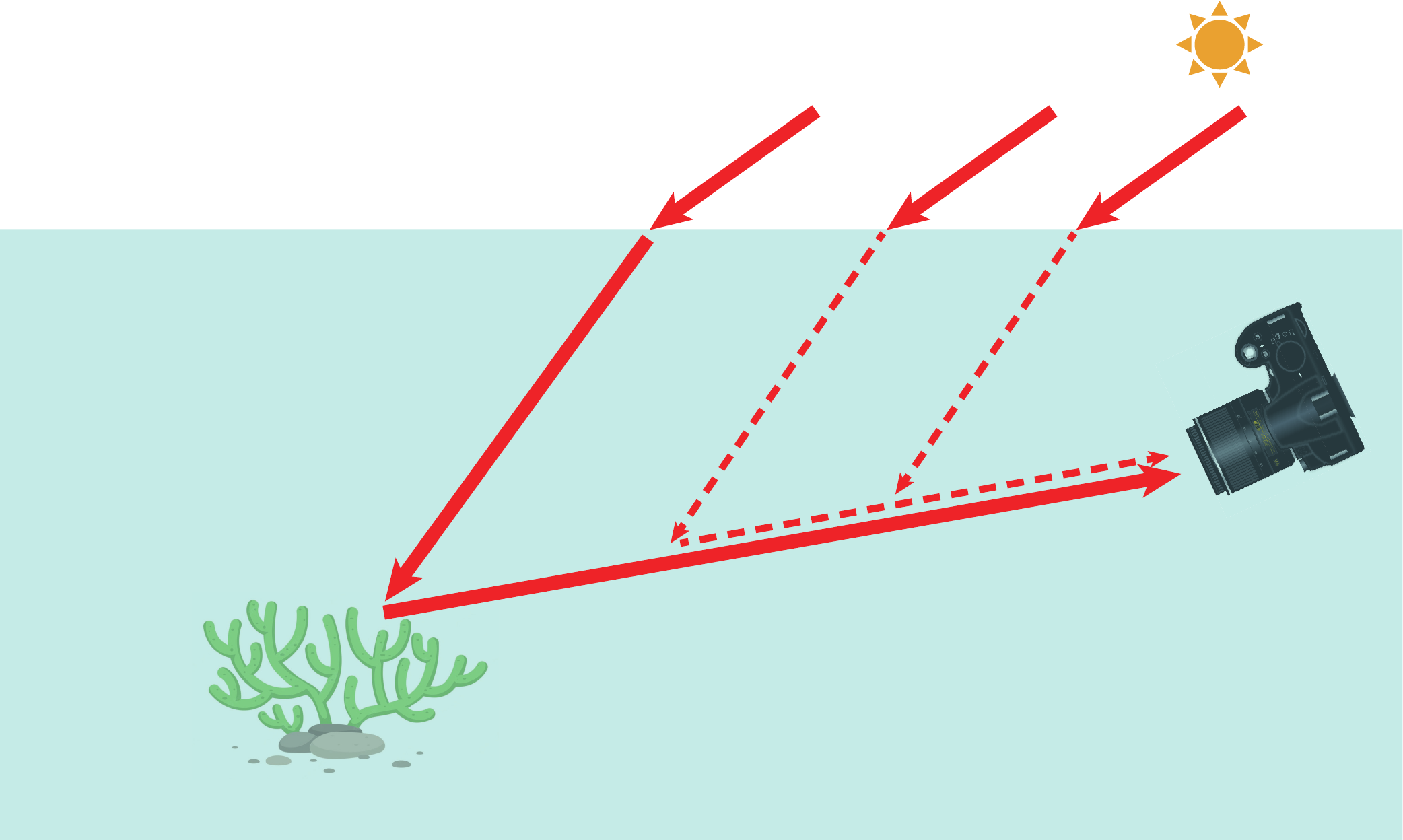}
\\ [1mm]
\includegraphics[width=0.8\linewidth]{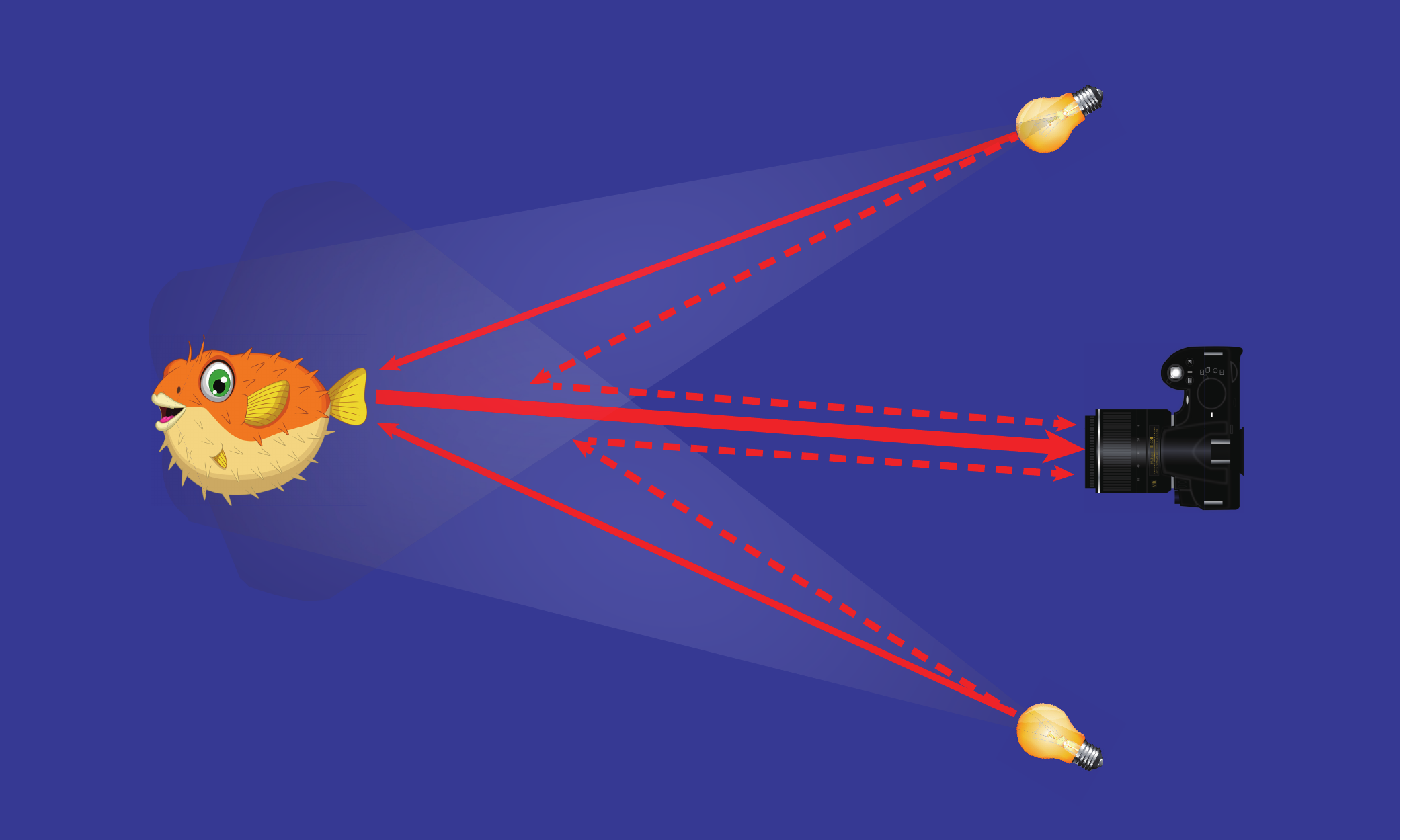}
\caption{Two popular underwater image formation models used in underwater image restoration. Top: Shallow water image formation with homogenous illumination from the sunlight. Bottom: Deep water image formation under artificial illumination. }
\label{fig_image_formations}
\end{figure}

When a specific underwater image formation model (or rendering pipeline) is predefined, the water optical parameters can be estimated from redundant observations, either from multi-view images or different parts in a single image. The image restoration can be considered as the inverse rendering procedure of the underwater images with estimated parameters. Two popular physical models are the Atmospheric Fog (AF) Model (Eqt. \ref{FogModel}) and the Jaffe-McGlamery (J-M) Model \cite{mcglamery1980computer, jaffe1990computer} (see Fig. \ref{fig_image_formations}). 

The AF Model (and its modifications) is widely used due to its simplicity. It assumes the scene is illuminated homogeneously and the total backscatter is defined by a background light (also named water color, veiling light \etal), which depends on the water. 
Each pixel in an underwater images is described as a weighted combination of the true color $I_0$ and the background light $B_\infty$, and the underwater color is interpolated from these two values. 
The weight on the true color term is the transmission $T$, which can either be directly estimated from priors or computed from the estimated attenuation parameter $\eta$ ($T=e^{-\eta d}$). %The attenuation only consider the path from the object to the camera as the attenuation from the water surface to the object is backed into the background color already. 
The weight on the background light term is often expressed as $1-T'$. Here, $T'$ can be equal to $T$, or computed according to another parameter $T'=e^{-\eta' d}$. 

The advantage of the AF Model is that it only contains a few parameters (no integral involved) and does not require many redundant observations from multiple images. The information extracted from a single image is sufficient to estimate these unknown parameters. 
However, the drawback of this model is also obvious: $B_\infty$ is not able to describe complex total backscatter patterns, especially for Type III and IV images \cite{song2021deepsea}. 
The J-M Model is a more complex underwater image formation model that addresses the manifold scattering pattern cased by artificial point light sources. It integrates the scattered light along the viewing ray from all light sources, taking into account the attenuation along the entire transmission from the light sources to the object and then reflected to the camera. Estimating water parameters using the J-M Model typically requires multi-view correspondences. 

When the underwater image formation model is defined, in principle, it is possible to estimate scene depth, water parameters and lighting configuration simultaneously from multi-view images. However, this problem is degenerate in practice, and often the lighting configuration is known in advance to estimate the other parameters. 
This is known as underwater photometric stereo problem  \cite{narasimhan2005structured,tsiotsios2014backscatter,murez2015photometric,tian2017depth,fujimura2018photometric}. 
Similar concept is also used in the restoration approach where the traditional image formation models are replaced by a Monte Carlo ray-tracing pipeline \cite{nakath2021in}. 
If the scene depth is known as well, the water parameters can be estimated directly and used to correct image color \cite{bryson2016true}. 
The J-M Model requires knowledge of each light source individually, limiting its feasibility under complex lighting conditions. 

To tackle unknown lighting pattern, subjective approaches based on qualitative criteria are often used.These include methods based on the illumination-reflectance model \cite{pizarro2003toward,johnson2017high,bodenmann2017generation,koser2021robustly}, histogram equalization \cite{eustice2002uwit,lu2013underwater} and homomorphic filtering \cite{singh1998quantitative,singh2007towards}. However, these methods primarily focus on correcting the lighting pattern to unify the brightness in the image, the color consistency with no guarantee of color consistency and proper removal of backscatterred signal (additive noise). Moreover, some of them assume a flat seafloor andconstant lighting pattern throughout image sequences, which is unsuitable for complex scene.

In our previous work \cite{song2021deepsea}, we pointed out that the backscatter pattern remains relatively stable within the viewing frustum in front of the camera. 
To accelerate the rendering procedure, a 3D lookup table was utilized to store the pre-rendered backscatter pattern. 
Building on this structure, this paper proposes a novel and versatile solution for underwater image restoration that addresses the limitations of existing methods such as AF, J-M, and qualitative criteria-based models. It excels in restoring the true colors of underwater images and effectively eliminates the uneven lighting artifacts induced by artificial light sources, which can handle illumination conditions ranging form simple to complex. The key contributions of our work are as follows: 
\begin{itemize}
\item We begin by categorizing various types of underwater image formations based on their illumination conditions and analyze their characteristics. A general underwater image formation model is the then presented with simple formulation but can effective address different types of underwater images. 
\item Based on the general model, we introduce a parameter-free restoration approach, which applies a 3D lookup table in front of the camera to robustly estimate and compensate water and lighting effects. Our proposed approach does not require additional knowledge of underwater environments like lighting conditions and water properties. It can not only restore the color of underwater image sequences, but also compensate the inhomogeneous lighting patterns caused by the artificial illumination. Furthermore, it preserve the uniform brightness and true color across image sequence, which is crucial for following 3D reconstruction and photo mosaicing process. 
\item We explore different constraints for estimating the parameters of the lookup table and systematically analyze the capacity of our method for restoring different types of images. The method is tested and evaluated on various datasets. Once the lookup table is estimated (calibrated), it can be used directly for image batch processing, which is particularly beneficial for large-scale data. 
\end{itemize}
%-------------------------------------------------------------------------
\section{Background principles}
This section describes the concept of using a 3D lookup table to describe the light and water effects in front of the camera and presents a general solution for restoring underwater image sequence under complex illumination. In order to estimate the parameters in the lookup table, several constraints are discussed in Section \ref{Constraints}. 

\subsection{Concept of underwater image formation and restoration}
\label{sec:general_model}

In considering the AF and J-M model, we assume that object shading has been compensated, these models can all be summarized by a combination of a multiplicative term (direct signal) and an additive term (backscattered signal):

\begin{equation}
\label{eqGeneralModel}
I =  \alpha \cdot I_0 + \beta. \quad \quad (\alpha, \beta > 0)
\end{equation}

In underwater images, pixel intensity for each channel $I$ is expressed as the product of the object abedo ${I_0}$ and the transmission factor ${\alpha}$, added by the backscatter component ${\beta}$. 
It is important to note that the intensity observation referred to in the following contents always refers to the intensity after shading compensation. Assuming that the object surface is Lambertian, shading compensation can be performed by dividing the original pixel intensity by $\cos \theta$, where $\theta$ is the angle between object surface normal and incoming light. 
We approximate the light originates from the camera position, and the surface normal can be calculated from the corresponding depth map. 
Underwater image restoration can be considered as an inverse processing that aims to recover the object abedo from the underwater observations $I$. It is achieved by subtracting ${\beta}$ from the observed image and dividing the result by ${\alpha}$: 

\begin{equation}
\label{eqGeneralModel_restore}
I_0 =  \frac{I-\beta}{\alpha}.
\end{equation}

\begin{figure}
\centering
\includegraphics[width=0.8\linewidth]{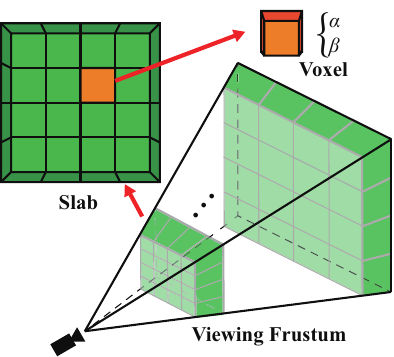}
\caption{Proposed 3D lookup table structure. The camera viewing frustum is sliced into several slabs and each slab is constructed by a plane of voxels. Each voxel with in a slab stores two parameters: a multiplicative factor $\alpha$ and a additive factor $\beta$, for each color channel. These parameters represent the combined effect of lighting and water at that particular 3D position. 
Giving the stable lighting and water conditions during a single mission, either under homogeneous illumination in shallow water or co-moving artificial light source in deep water, the parameters in the lookup table are relatively fixed, enabling rapid batch restoration of entire image sequences. }
\label{fig_3D_view_frustum}
\end{figure}

In \cite{song2021deepsea}, we introduced a novel approach for accelerating the backscatter rendering in underwater images. It involves slicing the 3D view frustum in front of the camera into multiple slabs, with each voxel in the slab storing a pre-computed backscatter value for each RGB channel. 
This allows for direct interpolation of the backscatter component for each pixel based on its 3D position in the local camera coordinate system.
This paper adapts the same structure as the parameters container and each voxel stores one multiplicative factor ${\alpha}$ and one backscatter factor ${\beta}$ for each color channel, forming up a lookup table (see Figure. \ref{fig_3D_view_frustum}). 
This model is not only suitable for underwater image applications, but can also be extend to in air cases such as in fog or with active illumination. 

\subsection{Observations and Constraints}
\label{Constraints}
Estimation of the parameters in the lookup table can be accomplished through a variety of constraints derived from underwater images. This paper introduces several physical constraints that can be leveraged, including Known Color Constraints, Correspondence Constraints, Smooth Constraints and Pure Water Constraints. These constraints are grounded in real-world physics and provide effective means for accurately estimating the lookup table parameters. 

\subsubsection{Known Color Constraints} \label{sec:KnownColorConstraints}
\begin{figure}
\centering
\includegraphics[width=0.5\linewidth]{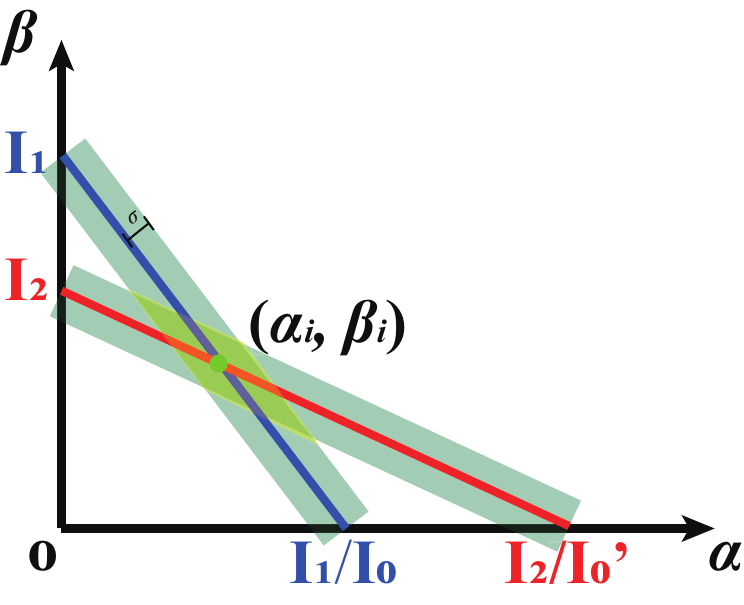}
\caption{One observed color ($I$) with a known color ($I_0$) can only provide a constraint on $\alpha$ and $\beta$ along a line in the $\alpha$-$\beta$ plane. 
To obtain a unique solution for each voxel, at least two observations with different known colors are required. 
As shown in the figure, the blue line is the constraint from one observed underwater color $I_1$ at voxel $V_i$ with known color $I_0$, while the red line refers to the constraint from another underwater color observation $I_2$ at the same voxel with second known color $I_0'$. 
The intersection point of the two lines (in green) provides the unique solution $(\alpha_i,\beta_i)$ for voxel $V_i$. 
Due to the uncertainty $\sigma$ in the observations, each line is only constrained in the green interval and the ambiguity of the solution is defined by the intersection of the two constraint regions (in yellow). 
To minimize this ambiguity and reduce the uncertainty of the solution, slopes of two lines ($-I_0$ and $-I_0'$, respectively) should be perpendicular to each other in order to achieve minimum intersection of intervals. Therefore, two known colors with widely disparate values should be used for the observations.}
\label{known_color_graph}
\end{figure}

When filming an object with known color (abedo), Eqt. \ref{eqGeneralModel} can be used directly to form the known color constraint, which becomes an equation of a simple line on the $\alpha$-$\beta$ plane. 
However, as shown in Fig. \ref{known_color_graph}, single known color constrain is insufficient to solve for the two unknown parameters in each voxel. At least two observations ($I_1$ and $I_2$) with different known color objects ($I_0$ and $I_0'$, respectively) on the same voxel $V_i$ are required to obtain the unique solution for corresponding $\alpha_i$ and $\beta_i$ (see Eqt. \ref{eqKnownColor}). 
Moreover, due to the errors in measurement, each known color constraint provides an interval of solutions rather than a single line. To minimize the intersection of intervals and reduce the uncertainty of the solution, the two known colors are supposed be widely disparate.  

\begin{equation}
\begin{cases}
\label{eqKnownColor}
I_1 = \alpha_i \cdot I_0 + \beta_i.\\
I_2 = \alpha_i \cdot I_0'; + \beta_i.\\
\end{cases} 
\end{equation}

In principle, an ideal diffuse object that reflects all visible light wavelengths equally and a perfect black body that absorbs all incoming light will minimize the uncertainty of the solutions. 
In this case, the backscatter factor ($\beta$) in the lookup table can be directly measured by filming the black body in the medium at different distances. Once all $\beta$ values are fixed, $\alpha$ values can be computed directly by subtracting the corresponding $\beta$ from images of the ideal diffuse object ($\alpha = (I-\beta) / I_0$, where $I_0 = 1$).

\subsubsection{Correspondence Constraints}

\begin{figure*}
\centering
\includegraphics[width=1.0\linewidth]{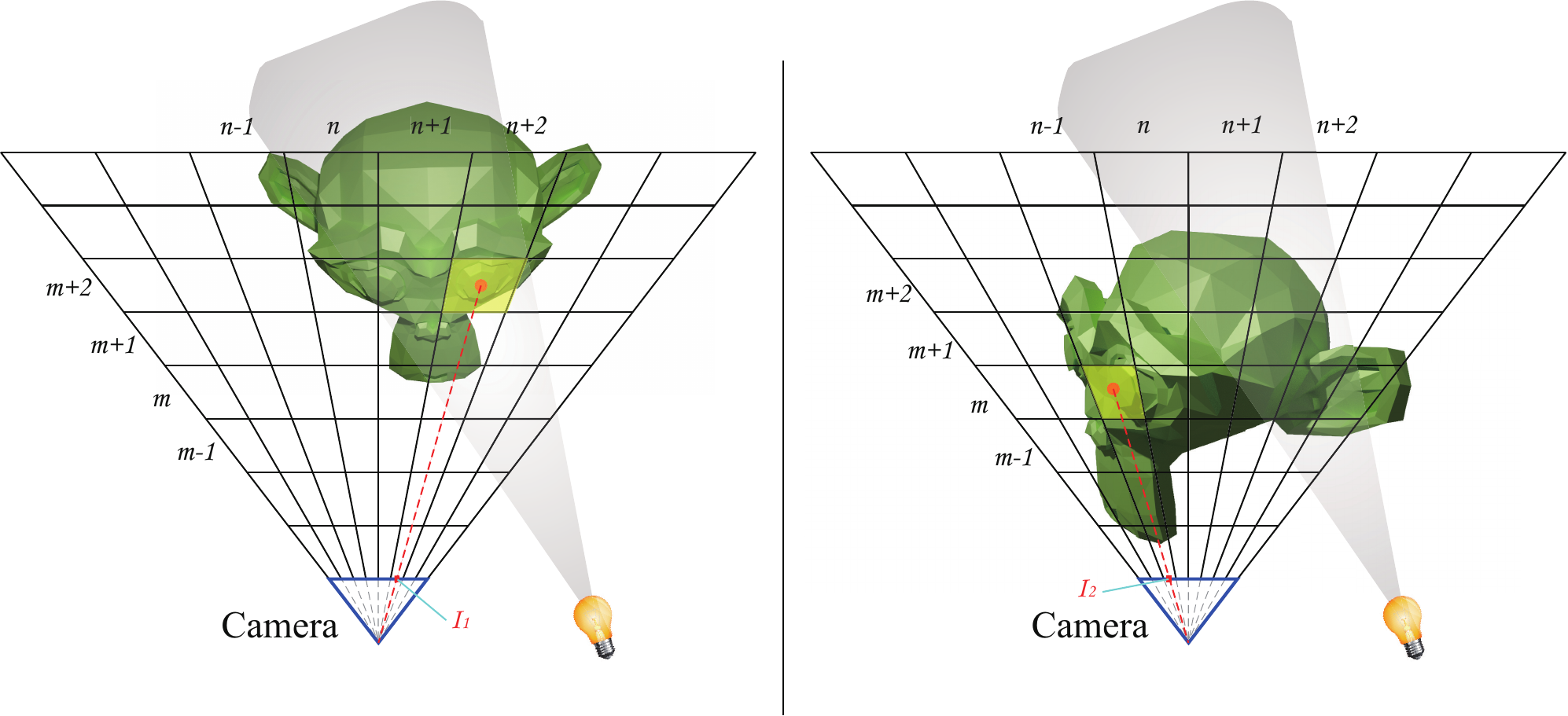}
\caption{Correspondence constraint can be constructed from image correspondence in the scene. As it is shown, a point is filmed by two images with pixel intensities $I_1$ and $I_2$, correspond to voxels $V(n+2,m+2)$ and $V(n-1,m)$, respectively. Parameters for these two voxels are integrated to form one correspondence constraint: $\alpha_{(n-1,m)} \cdot I_1 - \alpha_{(n-1,m)} \beta_{(n+2,m+2)} - \alpha_{(n+2,m+2)} \cdot I_2 + \alpha_{(n+2,m+2)} \beta_{(n-1,m)} = 0$.}
\label{correnpondence_graph}
\end{figure*}

Similar to the feature matching problem in structure from motion, pixel color correspondents between images can be established in order to estimate the parameters in the 3D lookup table (see Fig. \ref{correnpondence_graph}). When the same object is filmed by two images w.r.t different voxels in the lookup table, two equations can be generated according to Eqt. \ref{eqGeneralModel}: 

\begin{equation}
\begin{cases}
\label{eqCorrespondence}
I_1 = \alpha_1 \cdot I_c + \beta_1.\\
I_2 = \alpha_2 \cdot I_c + \beta_2.\\
\end{cases} 
\end{equation}
Where $I_1$ and $I_2$ are the two different observed color of the correspondents which share the same unknown object abedo $I_c$. 
This type of constraint is not sufficient to directly estimate the lookup table parameters, as each pair of image correspondences contains four unknowns. Eqt. \ref{eqCorrespondence} can be extended to include multiple observations of the same point in different images, but this does not help in solving the problem as more unknowns are added to the equation system. 

At least four pairs of images observe four different colordeuts objects at the same position in the local camera coordinate system, it is possible to achieve a unique solution. Unfortunately, it is difficult to obtain such complex constraints in practice. Therefore, this paper constructs the constraint for each pair of correspondences, which can be further derived to: 

\begin{equation}
\label{eqCorrespondence_int}
\alpha_2 \cdot I_1 - \alpha_2 \beta_1 - \alpha_1 \cdot I_2 + \alpha_1 \beta_2 = 0.
\end{equation}

Extracting reliable color correspondences between images is a critical task. Traditional image corresponds is achieved by using key points (e.g. SIFT\cite{lowe2004distinctive} and SURF\cite{bay2006surf} features), which are based on gradient features and are distributed on image corners or edges where significant changes in pixel intensities. These areas usually have unreliable and inaccurate color information due to the rapid changes in intensity. Color correspondents require to be extracted from homogeneous region. This paper utilizes super-pixel \cite{achanta2012slic} to segment the image into patches, where each patch exhibits relatively homogeneous color. Specifically, the color informationfor each patch is extracted from its center, which is then used to estimate the lookup table parameters. 

\subsubsection{Smooth Constraints}

Each voxel in the lookup table is not ensured to be assigned with observations from images, additional constraints are required to impose smoothness on the estimated parameters. The smooth constraint can be expressed in a simple form as follows: 

\begin{equation}
\label{eqSmooth}
w_{s,\alpha} \cdot (\alpha(x,y,z) - \alpha(x \pm 1, y \pm 1, z \pm 1)) = 0.
\end{equation}
Here lookup table parameter $\alpha$ at gird position $(x,y,z)$ is smoothed with its six neighbors. Similar constraint can be applied to $\beta$. 
The choise of weight $w_s$ in the smooth constraint is crucial as it is intended to balance neighbouring parameters while preserving the complex light pattern.
Typically, voxels located further away from the light sources have smoother illumination, so they are supposed to have stronger weights in the smooth constraint compared to the closer ones. 
More details regarding to weighting of the smooth constraints are discussed in Section \ref{secWeightsAndAccuracy}. 

In addition, it needs to be noted that each observation from images each observation from the images may not exactly correspond to the center of a voxel. 
To prevent the resulting estimations in the lookup table from being pixelated, each observation is assigned to interpolated parameters based on its neighboring voxels, rather than the parameters at its nearest neighbor. 
This results in increased smoothness through the estimated parameters. In this paper, trilinear interpolation with eight neighbors is used to interpolate the lookup table parameters for each observation. 
However, having a unique solution on one point is not sufficient to assign unique values to its neighboring voxels.
Therefore, it is necessary to ensure that at least eight points with unique solutions are presented in each group of eight neighboring voxels. 

\subsubsection{Pure Water (Complete Backscatter) Constraints}

\begin{figure}
\centering
\includegraphics[width=0.9\linewidth]{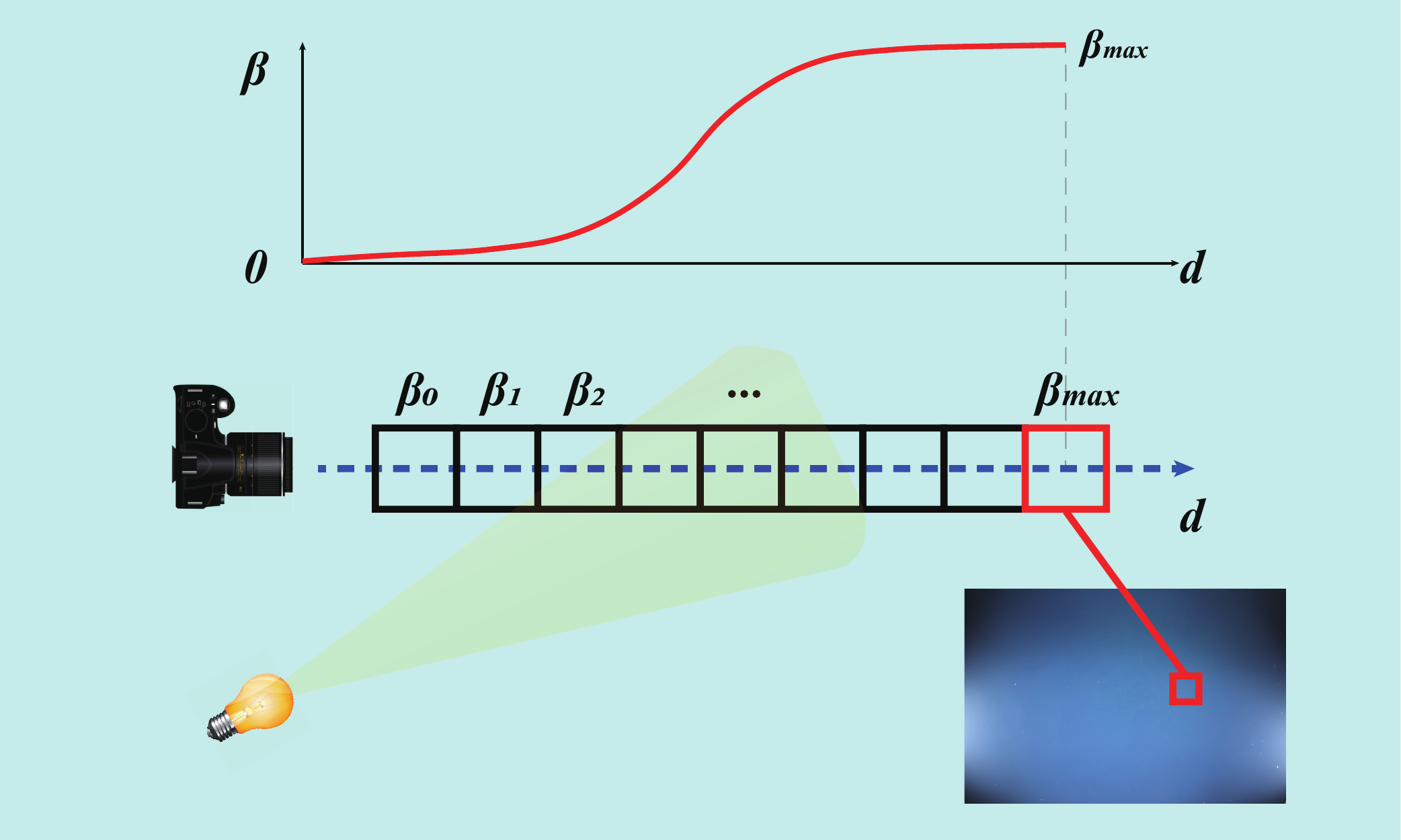}
\caption{The backscatter factor $\beta$ is monotonically increasing along each viewing ray. Pure water image records the full backscatter information present in the scene, which corresponds to the maximum $\beta$ value for each viewing ray.}
\label{fig_pure_water}
\end{figure}

During deep ocean missions, underwater vehicles take hours to dive down to the sea floor. During this period, camera records numerous images of pure water, containing only illuminated water in the scene. These images are usually considered as useless data for the mission. However, they contain the maximum illumination backscatter information, which can also be use to set up constraints for lookup table estimation (see Fig. \ref{fig_pure_water}). Each pixel in pure water image, denoted as $I_{pw}$, can contribute a direct constraint to all the $\beta$ terms at each slab $N$ along the same viewing ray:

\begin{equation}
\label{eqBS}
\beta_{N} \leqslant I_{pw}.
\end{equation}

This constraint establishes the upper bounds for the $\beta$ values. When underwater imaging platforms are operating at high attitude, pure water images can be directly used to subtract the backscatter component from the underwater images \cite{koser2021robustly}. %If pure water images are not available, a Fourier low pass filter can be used to extract the low frequency component from the minimum image throughout the image sequence. 
%(Proof of this assumption, show plots and rendering that first 5m slabs contains 95 percent of the BS information.) For single image, use blue channel to extract the bs pattern in blue channel and extend to other channels. 

\subsection{Hierarchical Parameter Estimation Strategy}
\label{HierarchicalStrategy}
Estimating parameters for the entire lookup table poses a challenge as it requires sufficient observations for each voxel to achieve a unique solution. 
To address this problem, this paper proposes a novel hierarchical strategy for parameter estimation that proceeds from coarse to fine resolution. 
The optimization solver starts to estimate the lookup table at very low resolution, and the estimated parameters as used as the initial values for the next iteration with higher resolution until the final target resolution is reached. 
This approach allows for a more efficient and accurate estimation of the parameters and enables us to fill the entire lookup table, even in areas where there are no observations available. 

\subsection{Weights and Accuracy}
\label{secWeightsAndAccuracy}

In the process of estimating the lookup table parameters, proper weighting of the constraints is crucial, as pixel observations may have varying degrees of uncertainty due to different distances and illumination conditions. 
To achieve this, a lookup table is pre-rendered under single point light illumination, using predefined water parameters, and is used to define the weights for the three types of constraints.

To compute the weights for the smoothness constraints, the mean gradients within and between the slabs of the lookup table are used. As a general trend, the illumination becomes weaker and smoother as the distance from the light source increases, resulting in parameter values that are closer in proximity. We calculate the mean gradient $\overline{\emph{grad}}$ of $\alpha$ and $\beta$ for each slab to measure its similarity, and use this to calculate the weights $w_S$ for the corresponding smooth constraints on slab $N$ and between neighboring slabs $(N, N+1)$. This is achieved through the following equations: 

\begin{flalign}
\begin{aligned}
& w_{s, \alpha} (N) = 0.01 \times 0.7 / \overline{\emph{grad}}_N \\
& w_{s, \alpha} (N, N+1) = 0.01 \times 0.7 / \overline{\emph{grad}}_{N,N+1} \\ 
& w_{s, \beta} (N) = 0.01 / \overline{\emph{grad}}_N   \\  
& w_{s, \beta} (N, N+1) = 0.01 / \overline{\emph{grad}}_{N,N+1} 
\end{aligned}
\end{flalign}
It is important to note that $\alpha$ and $\beta$ are in different value scales, and hence a factor of 0.7 which represents the average intensity of the scene, is included in $w_{s, \alpha}$ to bring them to the same scale. Additionally, an empirical value of 0.01 is used in all smooth weights to reduce their impact compared to other constraint types. 

Weights of observed pixel intensities for each color channel are determined by their signal-to-noise ratio (SNR).
The digital camera noise is usually categorized into three main sources: shot noise, dark current noise, and read noise. 
In underwater robotic mapping missions, fixed exposure time and a small aperture are often used to prevent motion blur and maintain a large depth of field. In such scenarios, the dark current noise portion in the image can be considered a constant term, and read noise is also constant as the entire image sequence is captured by the same camera and dynamic range. 
Pixel values have different uncertainties based on the scene depth and illumination conditions. Objects at further distances are usually under weaker illumination, leading to lower SNR and larger uncertainty due to fewer photons reaching the pixel. Additionally, forward scattering effect becomes more significant as the distance increases, which further degrades pixel observation quality. This effect can be modeled using a distance-dependent Gaussian point spread function (PSF) \cite{jaffe1990computer}. 
In this paper, we integrate the SNR and forward scattering models, along with the inverse distance weight, to calculate weights for pixel observations (known color and corresponding constraints) in the lookup table parameter estimation. The weight of know color constraints $w_{kc}$ is computed as follows:

\begin{equation}
\begin{aligned}
& w_{kc} = \frac{1}{d_v} \cdot \frac{snr}{(e^{0.5 * d})^2} \\ 
\text{where ~} & snr = I / (n_{shot} + n_{const})  \\
& n_{shot} = 0.01 \cdot \sqrt{\overline{mean}_N}  \\ 
& n_{const} = \overline{mean}_N / (snr_{0,g} * \overline{mean}_{0,g})
\end{aligned}
\end{equation}
Here, $\frac{1}{d_v}$ represents the inverse distance weight, and $d_v$ is the observed point's distance from the corresponding voxel center. The PSF is approximated by $\frac{1}{(e^{0.5 * d})^2}$, where $d$ is the camera distance to the observed point. 
The shot noise $n_{shot}$ is computed from the mean intensity $\overline{mean}_N$ of slab $N$, which can be approximated under the gray world assumption (with intensity 0.7) by $\overline{mean} = \overline{\alpha}_N \cdot 0.7 + \overline{\beta}_N$, where $\overline{\alpha}_N, \overline{\beta}_N$ are the mean values on slab N. 
The green channel of the first pre-rendered slab $snr_{0,g}$ is used as the reference value, which assumes a 20 db SNR. The constant noise for RGB channels $n_{const}$ can be computed by referring to the first slab green SNR. 

Similarly, the correspondence constrain weight $w_{c_{1,2}}$ can be computed from two corresponding known color weights according to the Pythagorean theorem: 

\begin{equation}
w_{c_{1,2}} = \frac{w_{kc_1} \cdot w_{kc_2}}{\sqrt{w_{kc_1}^2 + w_{kc_2}^2}}.
\end{equation}

\section{Experiments and Results on Lookup Table Parameters Estimation}
This section presents the method to estimate the lookup table parameters for image restoration under complex illumination conditions by utilizing the combination of the constraints mentioned above. 
It involves using known color calibration objects as references to estimate the lookup table parameters. 
Section \ref{sec:KnownColorConstraints} has discussed that the basic model (Eqt. \ref{eqGeneralModel}) contains two unknown parameters ($\alpha$ and $\beta$) for each color channel in each voxel of the 3D lookup table, at least two known colors on the calibration objects are necessary to estimate the parameters in each voxel. Meanwhile, the two known colors should be widely separated in order to obtain robust parameter estimations. 
%These methodologies focus on leveragingThe other constraints are employed as supplementary measures to fill the gaps left uncovered by the known color constraints. Additionally, we delve into the possibilities of primarily employing correspondences constraints to estimate the lookup table.

%Two distinct strategies are proposed to address the parameter estimation problem. The first strategy  The second strategy centers around utilizing correspondence constraints derived from image correspondences to estimate the lookup table.

To validate the effectiveness of the method, several experiments were conducted. 
The initial experiment involved the use of simulated in-air data with an artificial point light source to proof the concept of calibrating the lookup table using multi-view images and demonstrate its applicability in in-air applications. Subsequently, a real in-air lab experiment was conducted. 
The third experiment utilized simulated deep clear underwater datasets with two arbitrary color boards, followed by a simulated dataset with a turbid water setting, to test the effectiveness of our method. These experiments showcased that our approach is not limited to widely separated known colors and that the quality of restoration is closely related to the SNR of the input images. 
Furthermore, a real-world lab experiment was performed, employing a single chessboard with two color patches to demonstrate the possibility of simultaneous geometric calibration and lookup table estimation, which provides a practical solution for real-world applications. %Lastly, a real ocean experiment was conducted to illustrate that when the known color information is insufficient to cover the entire lookup table, the correspondence constraint can play a major role in estimating the lookup table. 
Furthermore, we explore the feasibility and prerequisites for solving the lookup table parameters estimation without known the color of calibration objects. We present the restoration results obtained from simulated in-air data with artificial illumination and outline the challenges arising when applying this strategy to underwater scenarios. 

%\subsection{Parameter Estimation from Known Color}
%The first scenario for calibrating the lookup table mainly uses known color objects, which is commonly employed when reference objects with known colors are available. 
%In Section \ref{sec:KnownColorConstraints}, we have discussed that the basic model (Eqt. \ref{eqGeneralModel}) contains two unknown parameters ($\alpha$ and $\beta$) for each color channel in each voxel of the 3D lookup table, at least two known colors are necessary to estimate the parameters in each voxel. Meanwhile, the two known colors should be widely separated in order to obtain robust parameter estimations. %This paper demonstrates the method by using simple color boards with different colors to calibrate the lookup table. 

\subsection{In-air Calibration by Using White Calibration Boards}
\textbf{\textit{Validation on simulated data}}:
As mentioned previously, our method can also be applied to correct artificial light patterns in images captured in-air. In this case, backscatter can be ignored (i.e., all $\beta$ values are set to 0), and only the transmission factor $\alpha$ in each voxel needs to be estimated. 
Therefore, one known color object is sufficient to calibrate the lookup table. 
Thirty in-air images of a simple white board with corresponding depth maps were simulated from different distances using Mitsuba3 \cite{Mitsuba3}, where a point light source was placed at the same position and moved along with the camera.
40$\times$30 sample points were extracted from each image to calibrate the parameters in the lookup table. 
Each sample point provided a known color constraint, together with the general smoothness constraint, allowed us to estimate the parameters using the Levenberg-Marquardt algorithm based on Ceres Solver \cite{ceres-solver}.

\begin{figure}
	\centering
	\includegraphics[width=0.9\linewidth]{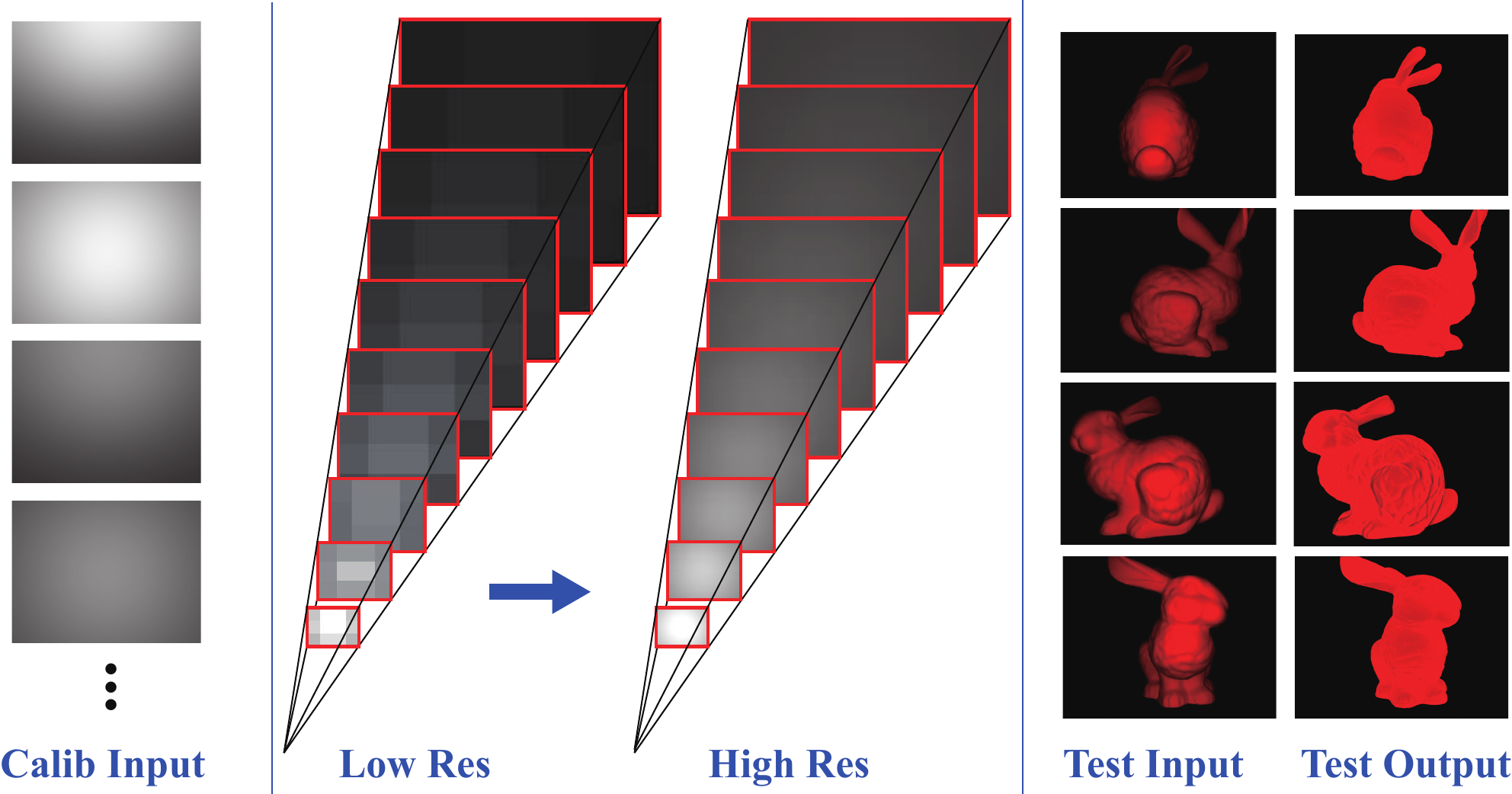}
	\caption{Experiment results on the synthetic in-air dataset. Left: simulated multi-view whiteboard images that were used as input to calibrate the lookup table. Middle: the coarse-to-fine estimation of the lookup table. Right: the test images under the same lighting configuration, along with their corresponding restored images after applying the estimated lookup table.}
	\label{in_air_calib_synthetic}
\end{figure}

As shown in Fig. \ref{in_air_calib_synthetic}, the coarse-to-fine strategy first estimated a low-resolution (4$\times$3$\times$10) lookup table, which was then used as the initial values for the later high-resolution (40$\times$30$\times$10) lookup table parameter estimation. 
Once the lookup table was estimated, we tested it on images of a uniform red color textured Stanford Bunny, which were simulated under the same lighting configuration. As can be seen, the proposed method effectively removes the uneven light pattern, resulting in properly recovered albedo of the model.

\begin{figure}
	\centering
	\includegraphics[width=0.4\linewidth]{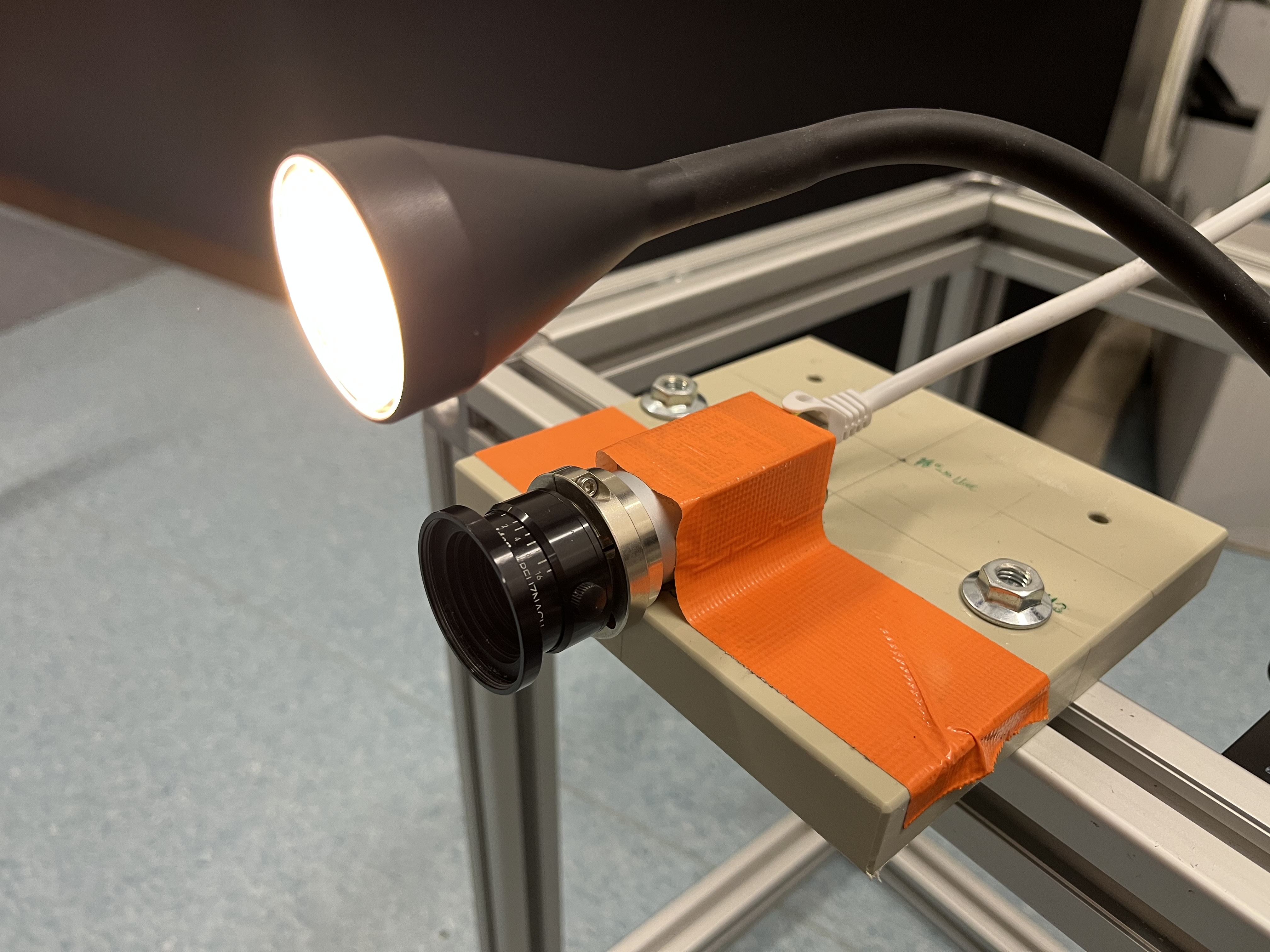}
	\includegraphics[width=0.4\linewidth]{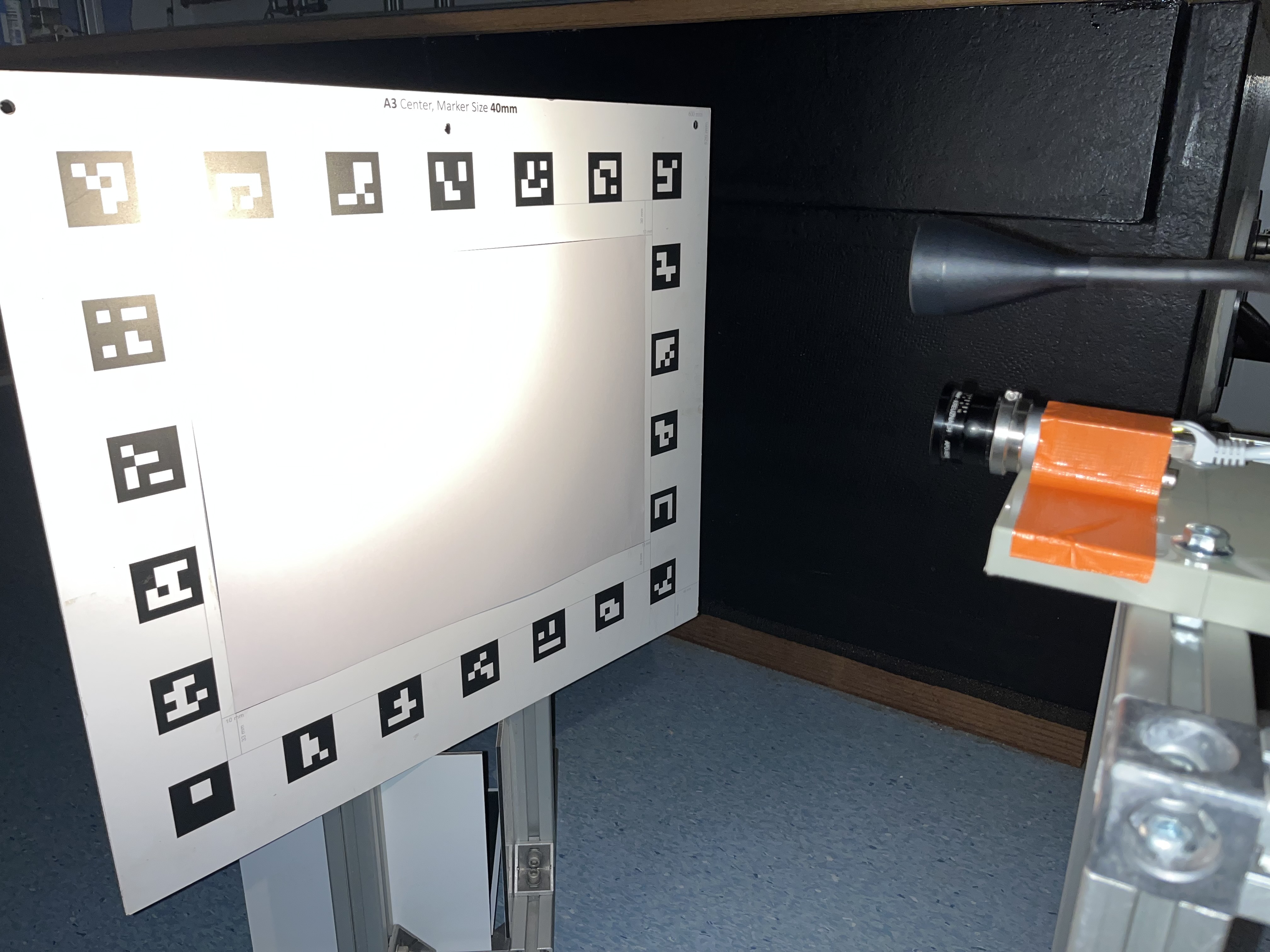}
	\caption{A rigidly-coupled camera-light system which was used in our laboratory experiment to capture several images of the self-designed calibration white board from different distances. Pixels in the center area of the board were used to calibrate the lookup table for the imaging system.}
	\label{in_air_calib_real_setup}
\end{figure}

\textbf{\textit{Validation on real experimental images}}:
A similar experiment was conducted on real captured images using a camera-light system (consisting of a Basler acA1920-50gm camera with a Schneider Apo-Xenoplan 2.0/20 lens and a normal lamp) that is rigidly-coupled (see Fig. \ref{in_air_calib_real_setup}). A self-designed calibration white board was used as the calibration object and multiple images of the board were captured from different distances to estimate the lookup table of the camera-light system.
To ensure accurate calibration, we assumed that the camera was already geometrically calibrated and that all captured images were undistorted accordingly. Additionally, we assumed that the camera's radiometric response was linear. 
The area of interest (AOI) was the center of the board covered with white Lambertian material. Sample points were selected from this area in the images to calibrate the lookup table. 
To provide depth information for the sampled points, AruCo markers on the board edges are detected and the relative poses between the camera and the board were estimated. 

\begin{figure}
	\centering
	\includegraphics[width=1.0\linewidth]{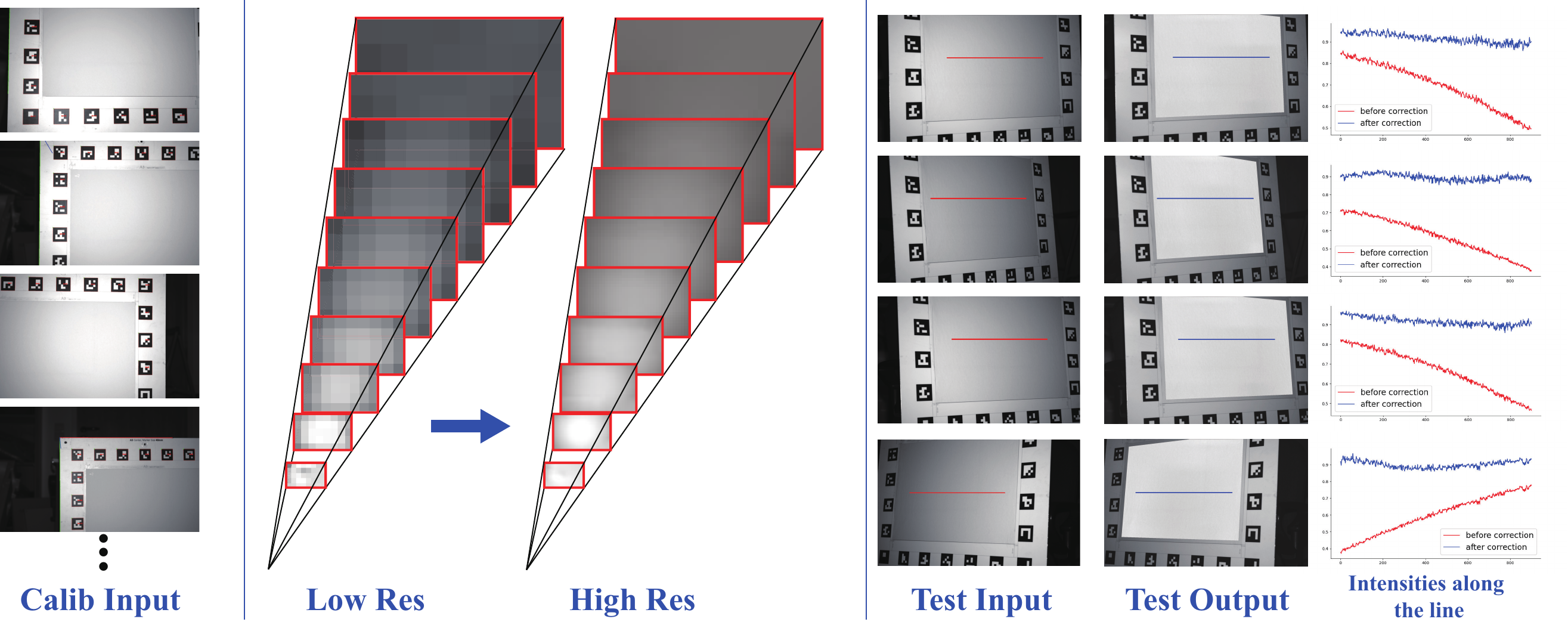}
	\caption{Experiment results on the real captured in-air dataset. 
	Left: Multiple images of the self-designed calibration white board are used to calibrate the lookup table. AruCo markers on the board are detected to estimate the poses of the board, providing depth information for the AOI. 
	Middle: Initial low-resolution lookup table estimation is refined to produce the final lookup table. Right: Test images of a tilted board captured under the same lighting configuration, with corresponding restored images.
	Pixel Intensities were sampled at the same position along a line from both test input and output images. Their plots indicate that the uneven lighting effect is properly removed after correction, resulting in a consistent abedo for the AOI.}
	\label{in_air_calib_real}
\end{figure}

Fig. \ref{in_air_calib_real} shows the results of the real in-air lab experiment. For estimating the lookup table parameters, sample points with computed depth were extracted in the AOI from thirty-five images of the calibration board. 
The coarse-to-fine approach (from 8$\times$5$\times$10 to 40$\times$25$\times$10) was used for calibration, and the final obtained high resolution lookup table was used to restore the test tilted board images captured under the same system. 
As shown in the figure, the correction process successfully removed the uneven light pattern. Moreover, the plotted intensity distributions along the lines in test images before and after the correction demonstrated that the recovered abedo over the entire AOI is relatively constant.

\subsection{Underwater Calibration by Using Two Different Color Boards}
In underwater cases, two unknown parameters ($\alpha$ and $\beta$) need to be estimated in each channel in each voxel, at least two known color objects are required to calibrate the lookup table. 

To validate the effectiveness of our proposed method, two underwater datasets with significantly different water types were simulated: clear deep water (Jerlov water type IA) and turbid coast water (Jerlov water type IC), using the state-of-the-art Monte Carlo ray-tracing technique based on Mitsuba3. 
Both datasets were rendered under the same camera-lighting setup, with a camera having 90 degree field of view and two rigidly co-moving point lights placed 40 cm to the left and right of the camera.
To test the robustness of our method, additional challenges were deliberately introduced to the simulated data. These challenges included limiting the number of samples per pixel (spp) to 512 during the Monte Carlo ray-tracing procedure, which resulted in an approximate 10\% error rate, and saving the simulated data as 8-bit RGB images rather than high dynamic range images. This decreased the accuracy and the SNR of the calibration data. 
Furthermore, two arbitrary color boards are used for simulating the calibration dataset (specifically, boards with RGB colors of [181, 110, 30] and [80, 160, 90]), instead of using widely separated colors like black and white. 
To account for the different visibility conditions in the two types of water, viewing frustums in different ranges were defined for each dataset. In the case of the clear deep water dataset, the lookup table was defined for depths ranging from 0.5m to 2.5m. For the turbid coast water dataset, the lookup table was defined for depths ranging from 0.5m to 1.5m, as beyond this point the object was no longer visible. 

\begin{figure}
	\centering
	\includegraphics[width=1.0\linewidth]{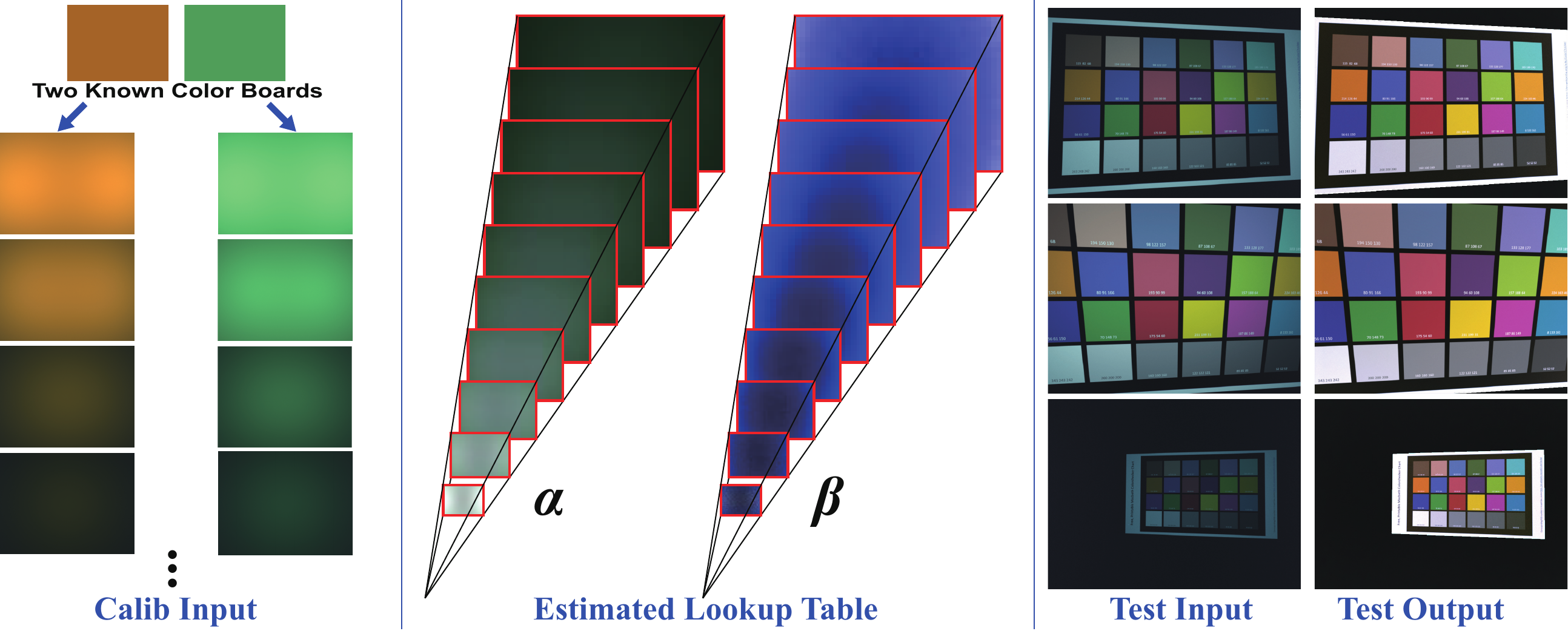}
	\caption{Experiment results on the synthetic clear deep water dataset. 
		Left: Input images of two known color boards used to calibrate the lookup table. 
		Middle: The final estimated lookup table visualizing the values of transmission ($\alpha$) and backscatter ($\beta$) parameters in the viewing frustum. The color mapping in the figure is scaled for better visualization. 
		Right: Test images of a color checker rendered under the same lighting and deep water conditions, along with the corresponding restored images obtained using the calibrated lookup table. }
	\label{uw_calib_clear}
\end{figure}

To calibrate the lookup table under deep water settings, thirty-one color board images with depth maps were simulated from different distances ranging from 0.5m to 2.5m. 
During the coarse-to-fine optimization, the unknown parameters $\alpha$ and $\beta$ in each voxel were estimated simultaneously. Fig. \ref{uw_calib_clear} illustrates the final obtained lookup table, which was used to restore the test images. The test images were generated from a virtual color checker that under the same environment settings as the calibration images. 

\begin{figure}
	\centering
	\includegraphics[width=1.0\linewidth]{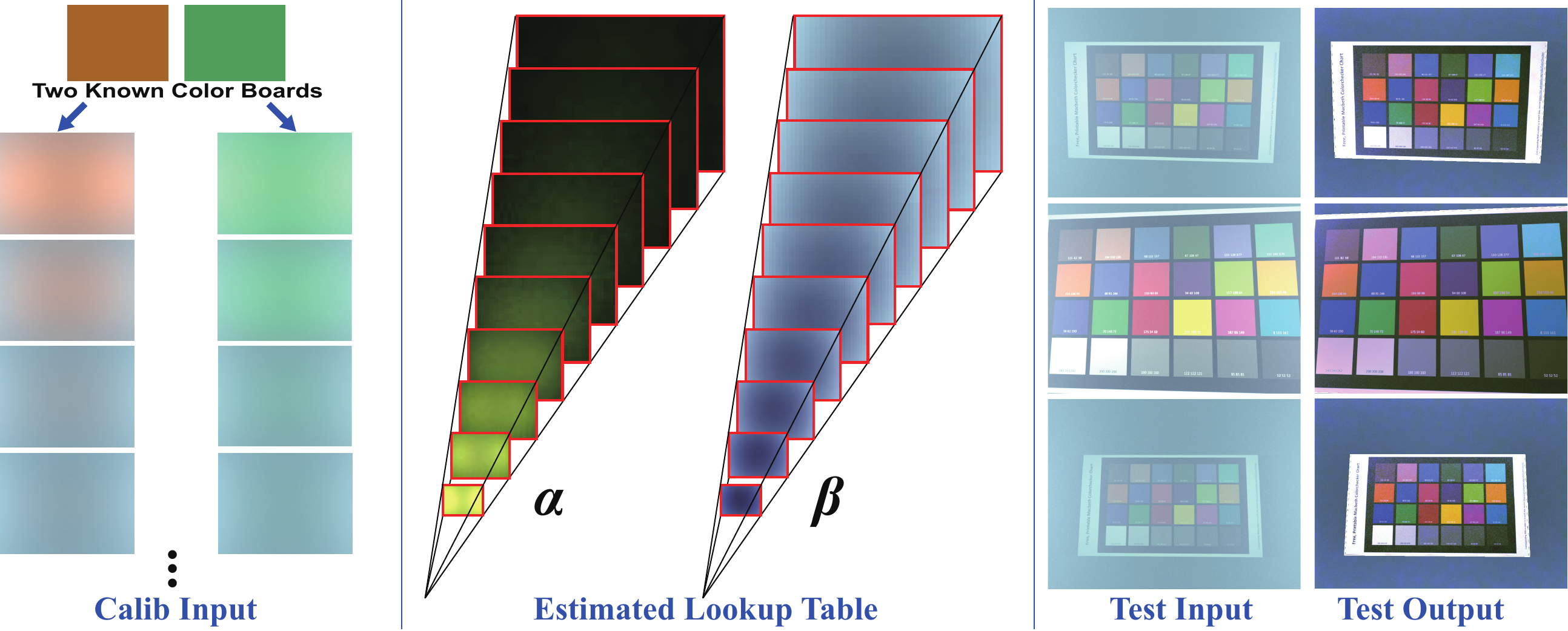}
	\caption{Experiment results on the synthetic turbid coast water dataset. 
		Left: Input images of two known color boards used to calibrate the lookup table. The images demonstrate the strong scattering effects present in the turbid coast water environment, resulting in poor visibility of objects.
		Middle: The final estimated lookup table showing the values of transmission ($\alpha$) and backscatter ($\beta$) parameters in the viewing frustum. The color mapping in the figure is scaled for better visualization. 
		Right: Test images of a color checker rendered under the same lighting and turbid water settings, along with the corresponding restored images acquired using the estimated lookup table. }
	\label{uw_calib_turbid}
\end{figure}

Similarly, in the simulated turbid coast water experiment, ten images for each color board at distances ranging from 0.5m to 1.5m were rendered to calibrate the lookup table. Once the lookup table was estimated, images of a virtual color checker under the same turbid water conditions were rendered to test the restoration method, as shown in Fig. \ref{uw_calib_turbid}. 

\begin{figure}
	\centering
	\includegraphics[width=1.0\linewidth]{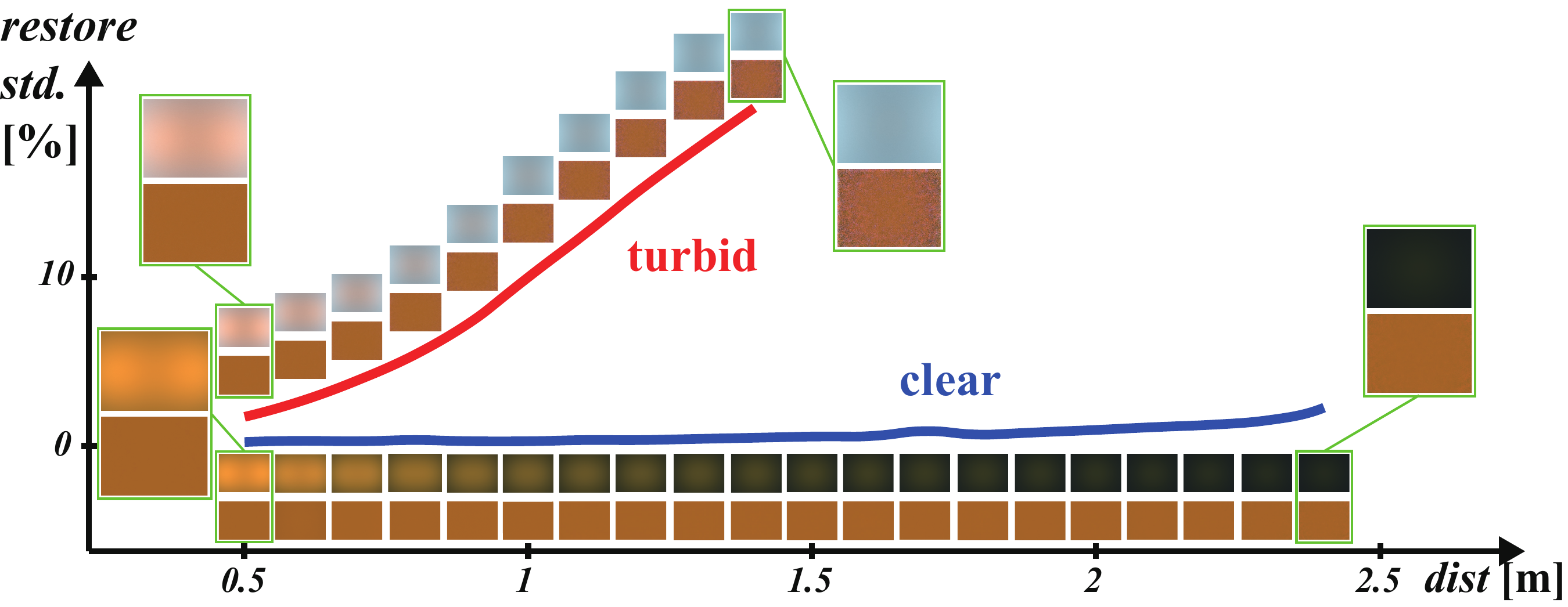}
	\caption{Standard deviation of restored images at different distances relative to the camera in the clear and turbid water datasets. The images displayed along each line represent the simulated underwater images, while the images below them depict the corresponding restored images used for computing the std. It is clear from the visualization that the SNR of turbid water images decreases much faster than that of clear water images, leading to higher std values in the restored images. }
	\label{uw_compare}
\end{figure}

As depicted in Fig. \ref{uw_calib_clear} and \ref{uw_calib_turbid}, our method effectively eliminates water and lighting effects while accurately restoring object albedo. The restoration quality is directly influenced by the SNR of the input images. In Section \ref{secWeightsAndAccuracy}, we discussed how images captured under stronger water effects and greater scene distances tend to exhibit lower SNR. 
In underwater imaging, as the scene distance increases, more light is absorbed by the water, leading to greater color attenuation, stronger forward-scattering effects and increased backscatter. 
Multiple images of a known color board at various distances relative to the camera in both clear and turbid water environments were rendered. These images were then restored using the corresponding estimated lookup table. The line plots shown in Fig. \ref{uw_compare} illustrate the standard deviation (std) of the restored images at different distances for both water conditions. As expected, the SNR of the images decreases with increasing distance, resulting in an increase in the std values of the restored images along the distance axis. In turbid water, the SNR decreases at a much faster rate compared to clear water images. This difference in SNR reduction leads to higher and more rapidly increasing std values in the restored images of turbid water conditions. 

Table \ref{tab:color_checker_measurement} presents the pairwise error of each color checker patch, computed as the absolute differences between the restored image and the ground truth color of each patch. 
In the clear deep dataset, the restored images exhibit high quality, with restoration errors mostly below the level of image noise. Despite the challenging conditions of the turbid coast dataset, characterized by poor visibility and very low SNR, some patches are even overexposed which  , our method still provides a significant visual improvement after restoration, with the majority of patch errors kept below 25\%. 

\begin{table*}
	\caption{Pairwise error (in \%) in RGB channels of each color checker patch, computed between the restored image to the ground truth values for the first test image in Fig. \ref{uw_calib_clear} and \ref{uw_calib_turbid}.}
	\label{tab:color_checker_measurement}
	\centering
	\resizebox{\textwidth}{!}
	{\begin{tabular}{lllllll}
		\hline\noalign{\smallskip}
		Clear deep & Col 1 & Col 2 & Col 3 & Col 4 & Col 5 & Col 6 \\
		\noalign{\smallskip}\hline\noalign{\smallskip}
		Row 1 & [2.11, 0.73, 2.67] & [0.61, 1.39, 1.60] & [0.86, 0.21, 4.42] & [0.03, 1.21, 0.26] & [2.18, 2.38, 10.67] & [2.15, 4.45, 14.58] \\
		Row 2 & [3.68, 2.79, 2.49] & [1.22, 0.81, 2.60] & [0.40, 0.06, 0.15] & [0.47, 0.84, 2.12] & [3.07, 3.14, 0.85] & [6.14, 3.44, 1.22] \\
		Row 3 & [0.96, 0.95, 0.12] & [1.23, 2.37, 2.25] & [1.00, 0.67, 1.38] & [3.11, 1.27, 0.48] & [3.35 1.33, 6.24] & [9.68, 2.90, 10.79] \\
		Row 4 & [4.37, 5.04, 5.08] & [1.52, 1.02, 8.74] & [9.25, 9.26, 6.75] & [0.06, 0.18, 2.35] & [7.89, 8.43, 9.43] & [7.82, 8.40, 6.95] \\
		\noalign{\smallskip}\hline
	\end{tabular}}

	\resizebox{\textwidth}{!}
	{\begin{tabular}{lllllll}
		\hline\noalign{\smallskip}
		Turbid coast & Col 1 & Col 2 & Col 3 & Col 4 & Col 5 & Col 6 \\
		\noalign{\smallskip}\hline\noalign{\smallskip}
		Row 1 & [1.01, 7.55, 14.53] & [1.66, 0.55, 22.65] & [1.56, 0.85, 25.76] & [1.89, 1.16, 3.01] & [7.19, 4.16, 23.64] & [7.63, 6.74, 17.29] \\
		Row 2 & [8.44, 3.76, 20.02] & [3.87, 5.21, 34.33] & [1.42, 3.86, 6.63] & [1.31, 7.36, 5.82] & [5.08, 2.89, 3.39] & [1.08, 5.46, 7.61] \\
		Row 3 & [12.40, 14.85, 39.10] & [7.10, 2.53, 14.96] & [0.18, 12.39, 7.22] & [5.30, 0.90, 8.24] & [2.04,  3.02, 21.46] & [18.97, 3.00, 25.33] \\
		Row 4 & [4.69, 4.69, 5.08] & [12.20, 12.23, 21.48] & [7.89, 5.92, 21.99] & [0.85, 1.64, 18.50] & [6.27, 9.54, 16.69] & [9.19, 13.14, 8.10] \\
		\noalign{\smallskip}\hline
	\end{tabular}}
\end{table*}

\subsection{Underwater Calibration by Using Single Board with Two Known Colors}

A more practical approach for obtaining two known colors involves distributing them on a single board, such as a chessboard with black and white patches. This allows us to perform the lookup table calibration by filming only a single board. 
Additionally, using a chessboard offers the advantage of simultaneous camera geometrical calibration, which is particularly beneficial for real robotic missions with limited operation time and energy supply. 

\begin{figure}
	\centering
	\includegraphics[width=0.436\linewidth]{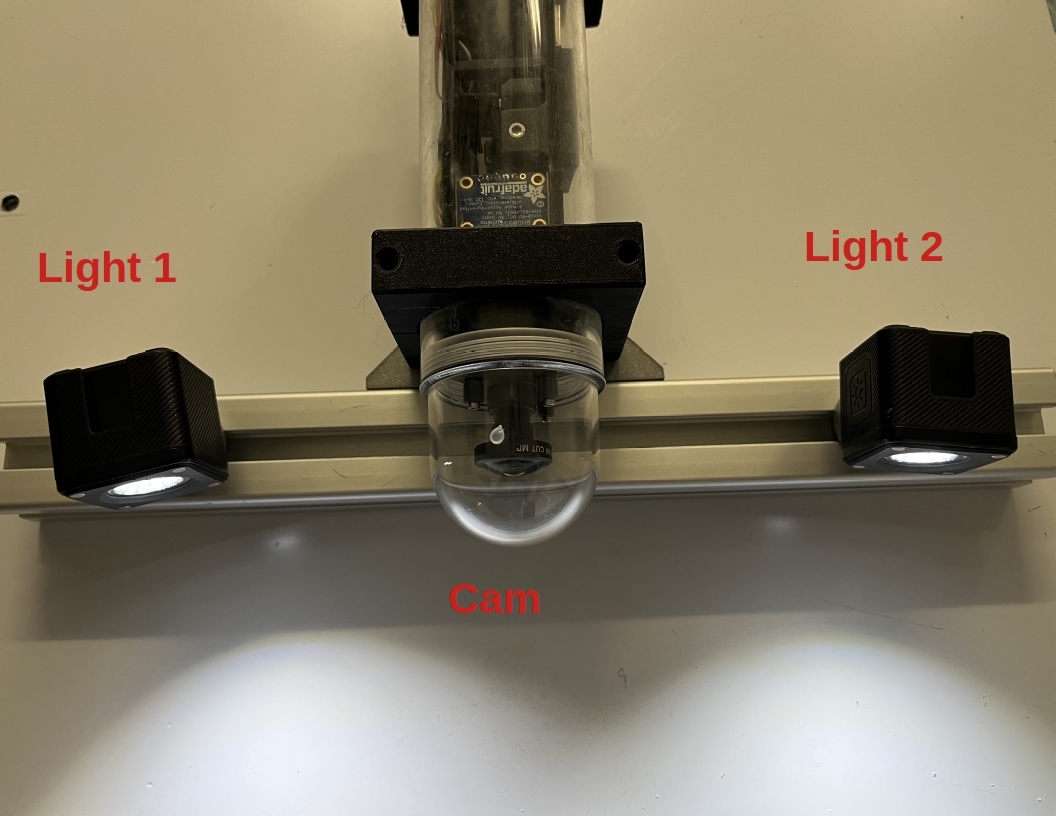}
	\includegraphics[width=0.45\linewidth]{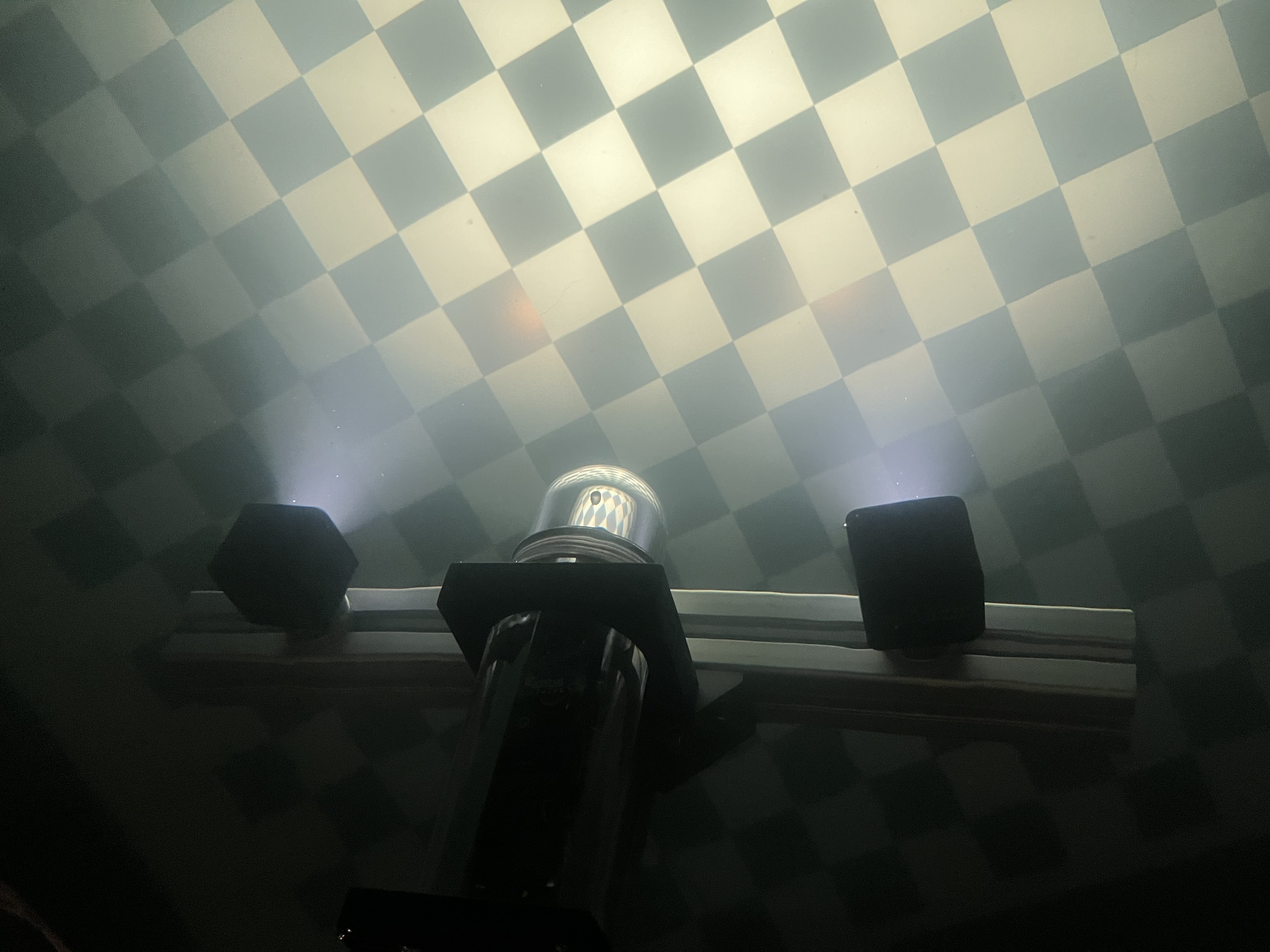}
	\caption{Left: The underwater camera system with dome port housing, accompanied by two rigid co-moving light sources. Right: A normal chessboard served as the calibration target and the center area of each chessboard patch was selected to facilitate the lookup table parameter estimation. }
	\label{uw_real_system}
\end{figure}

In our experiment, a custom underwater camera system enclosed in a dome-port waterproof housing (see Fig. \ref{uw_real_system}) was utilized. 
The system consisted of a Basler daA1600-60uc color camera equipped with an Evetar M118B029528W fisheye lens. 
Two rigidly co-moving light sources were positioned on the left and right sides of the camera, with a distance of approximately 15 cm from the camera. The camera was carefully adjusted to the center of the dome port using the techniques outlined in \cite{she2019adjustment} to eliminated the underwater refraction effect. Similar to the previous experiments, the camera underwent both geometric and radiometric pre-calibration. Additional materials were added into the water tank to augment the water effects, thereby intensifying the challenge for image restoration. For calibration, a standard chessboard was used as the target. Sample points were selected from the central region of each chessboard patch to calibrate the lookup table, and the relative poses between the camera and the board were estimated based on the chessboard corners, which were used to compute the depth information for each sample point. 

\begin{figure}
	\centering
	\includegraphics[width=1.0\linewidth]{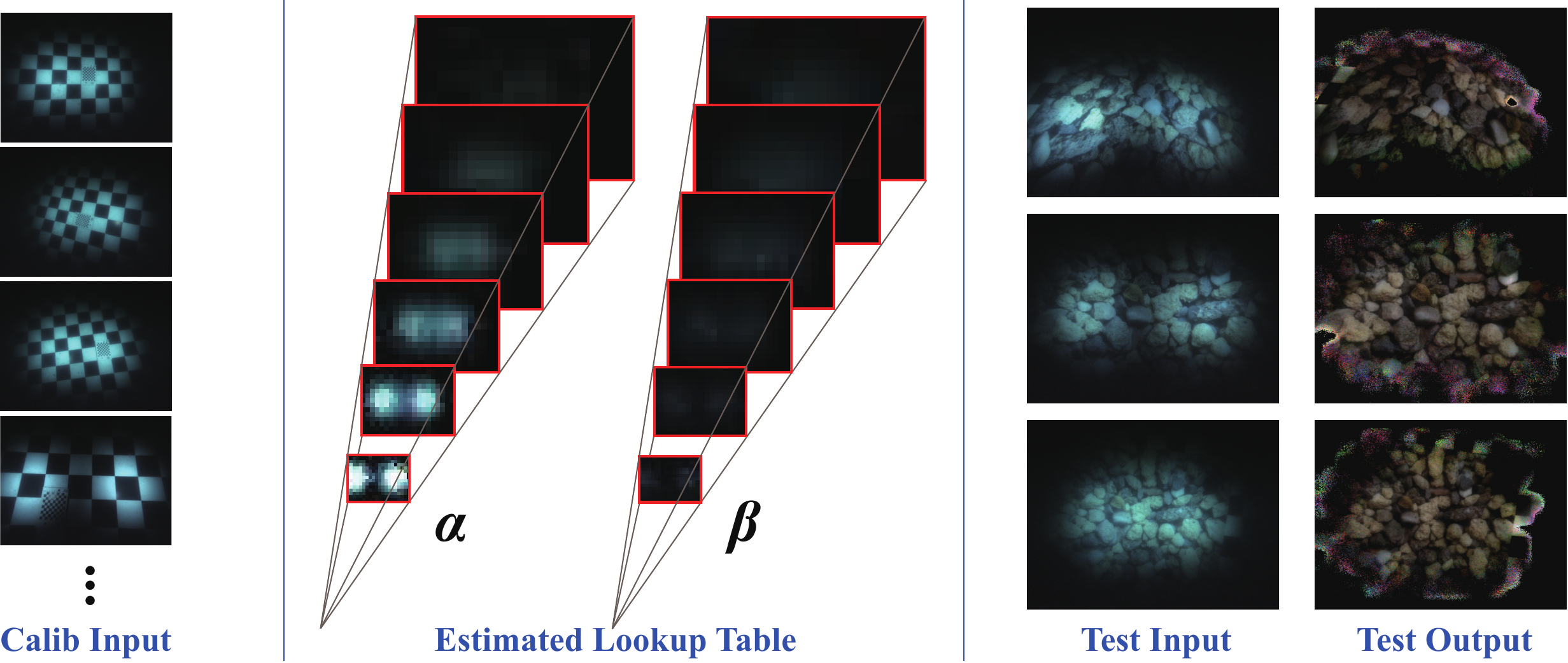}
	\caption{Experiment results on the real lab underwater dataset. 
		Left: Forty-one input images of a single chessboard were utilized for the calibration of the lookup table. These images exhibit noticeable light patterns and strong water effects, resulting in poor visibility. 
		Middle: The final estimated lookup table displaying the values of the transmission ($\alpha$) and backscatter ($\beta$) parameters within the viewing frustum. 
		Right: Test images captured by the same system under identical water conditions, alongside the corresponding restored images obtained using the estimated lookup table. The presence of colorful boundaries in the restored images can be attributed to the lack of informative data in those areas during the calibration process, leading to erroneous estimation of lookup table parameters. Moreover, the dark region exhibits a notably low SNR, further exacerbating the noise in these area.} 
	\label{uw_calib_tank}
\end{figure}

\begin{figure}
	\centering
	\includegraphics[width=0.3\linewidth]{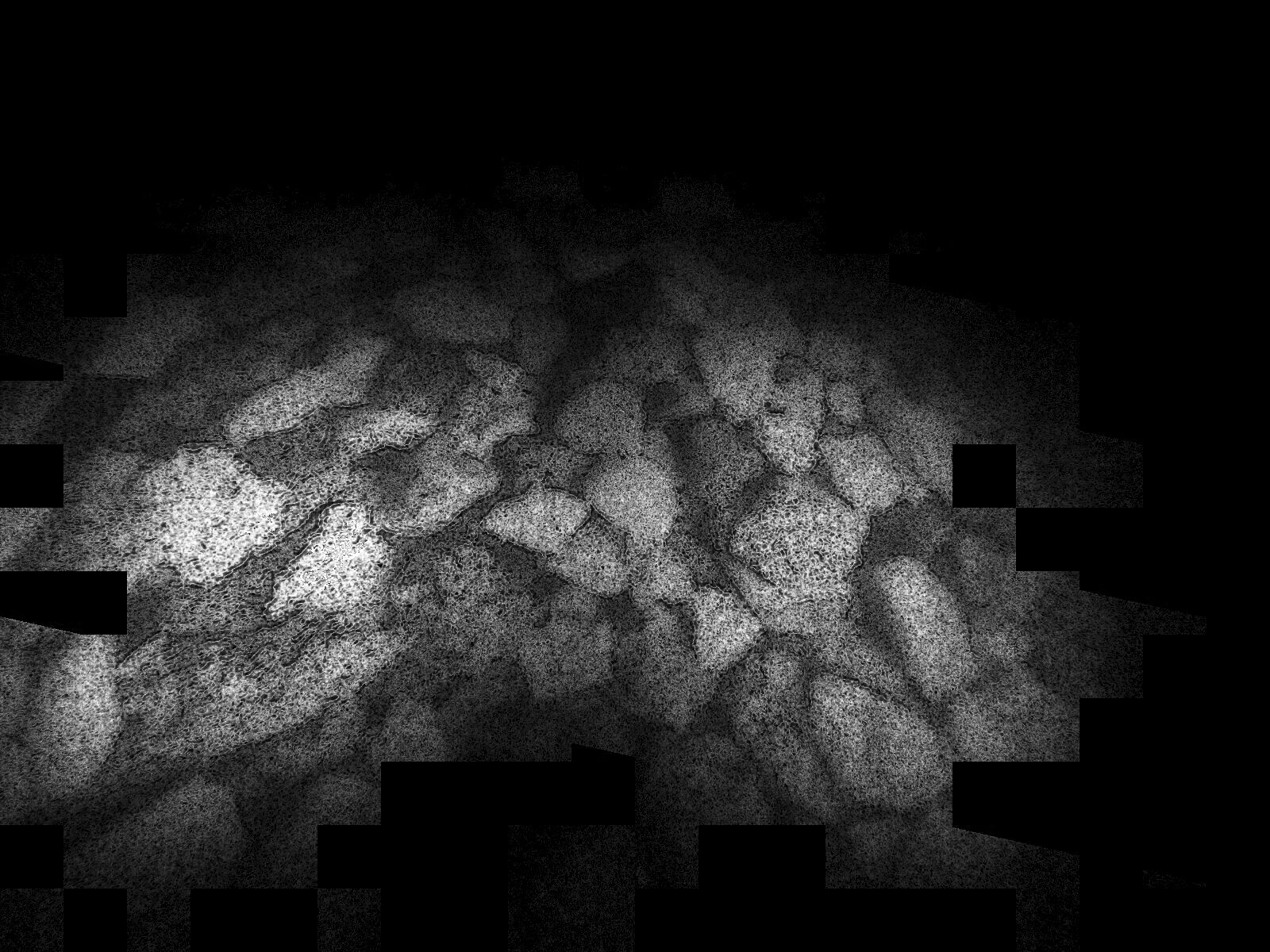}
	\includegraphics[width=0.3\linewidth]{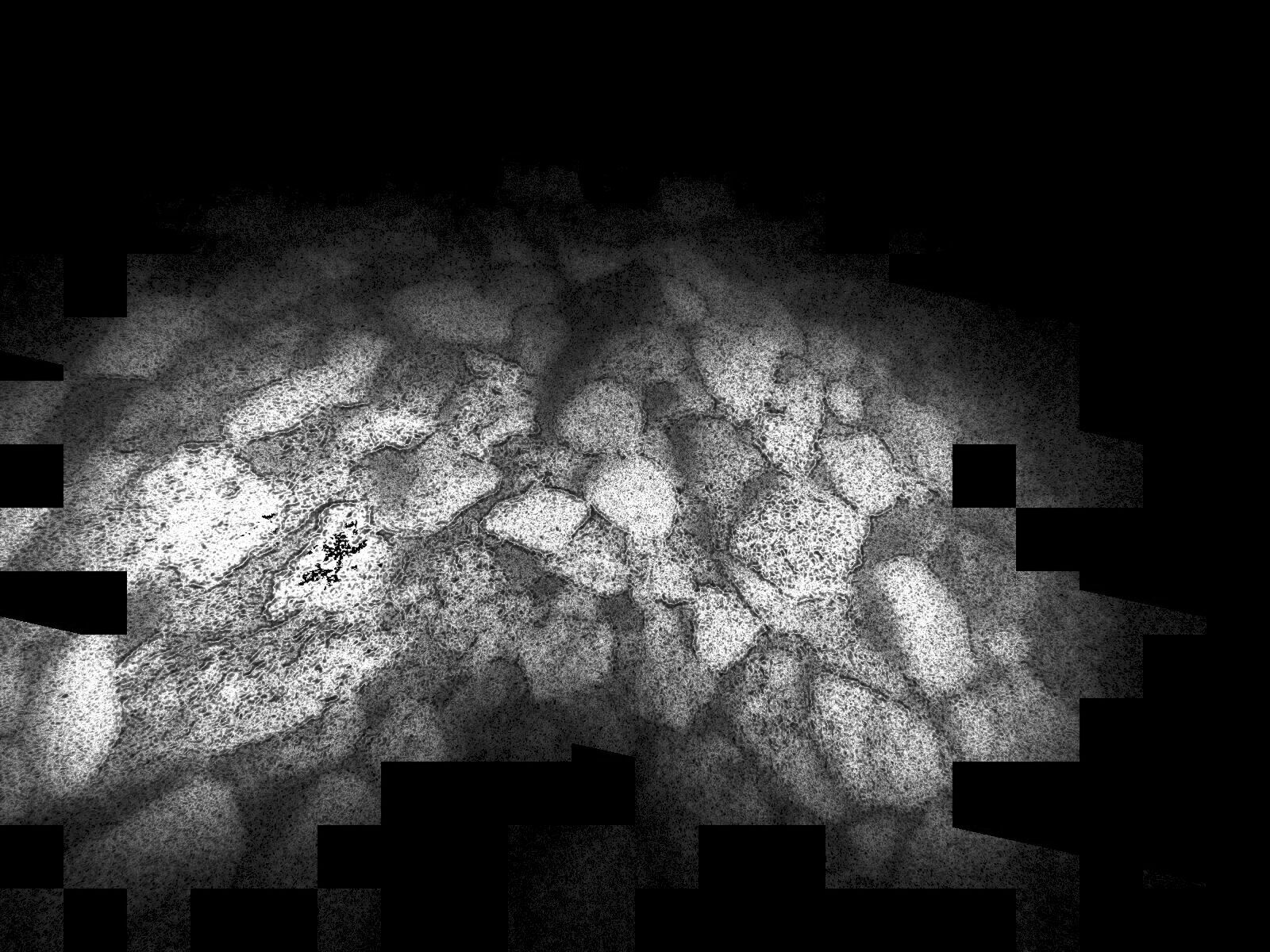}
	\includegraphics[width=0.3\linewidth]{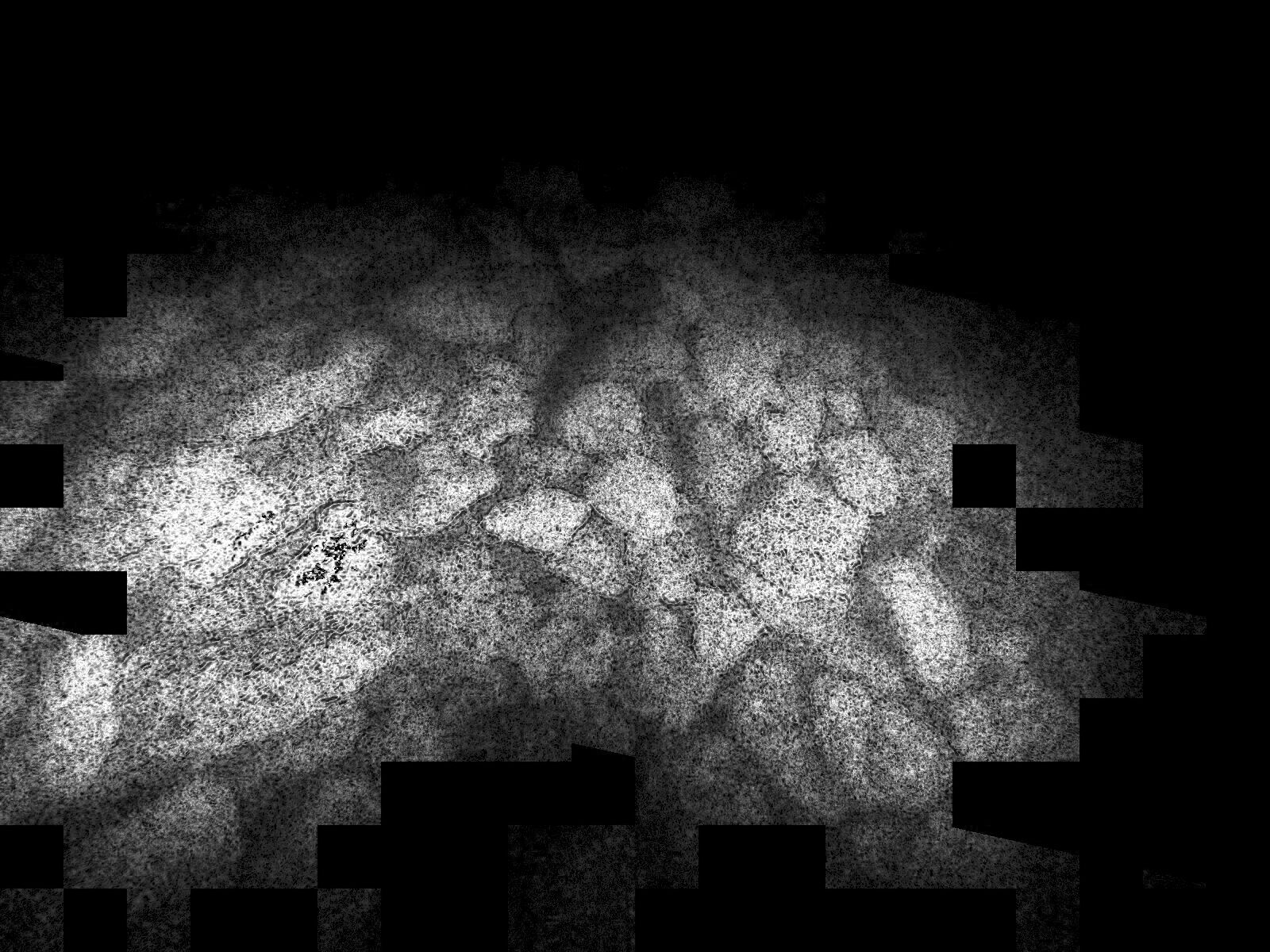}\\
	\includegraphics[width=0.3\linewidth]{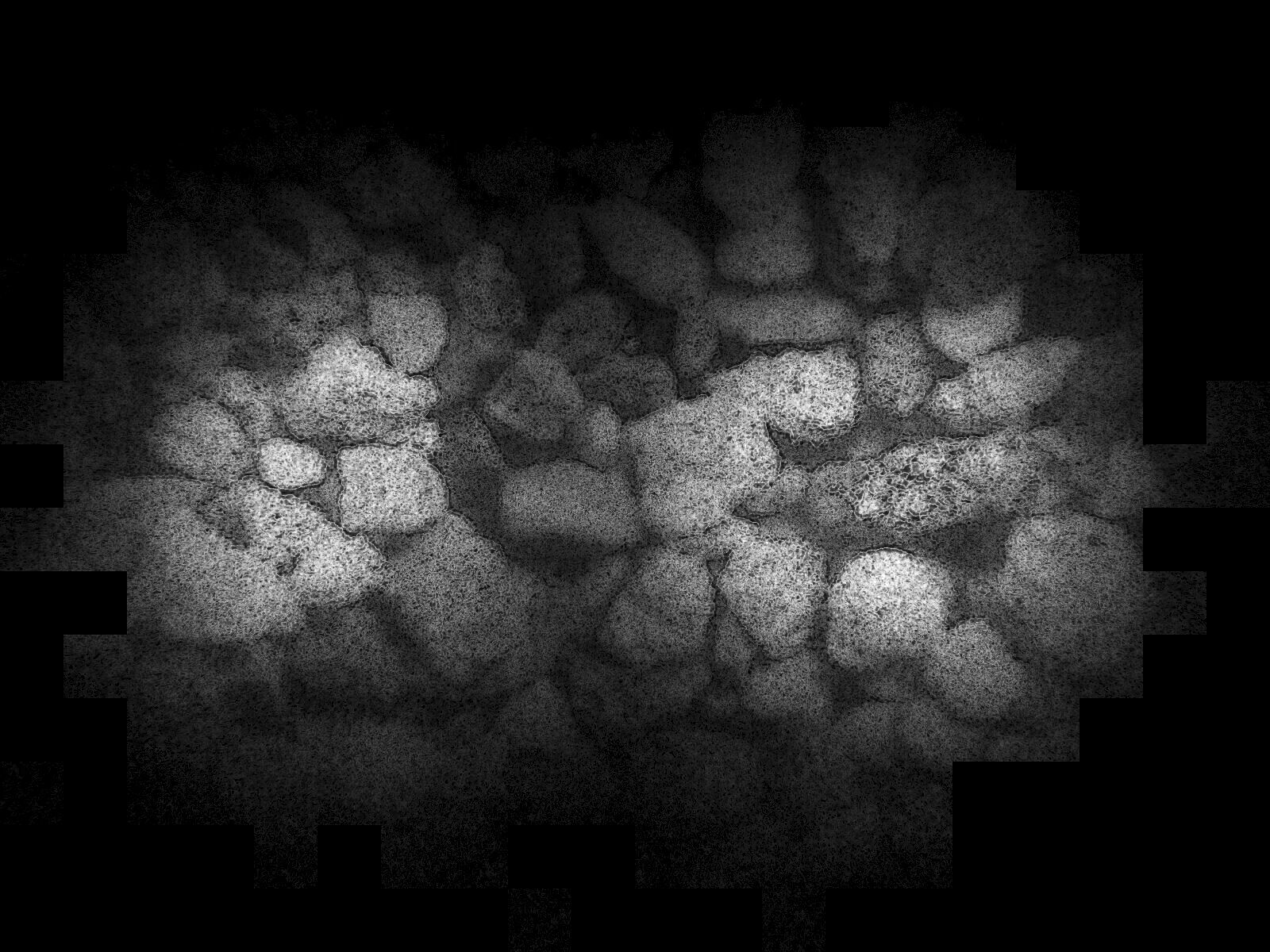}
	\includegraphics[width=0.3\linewidth]{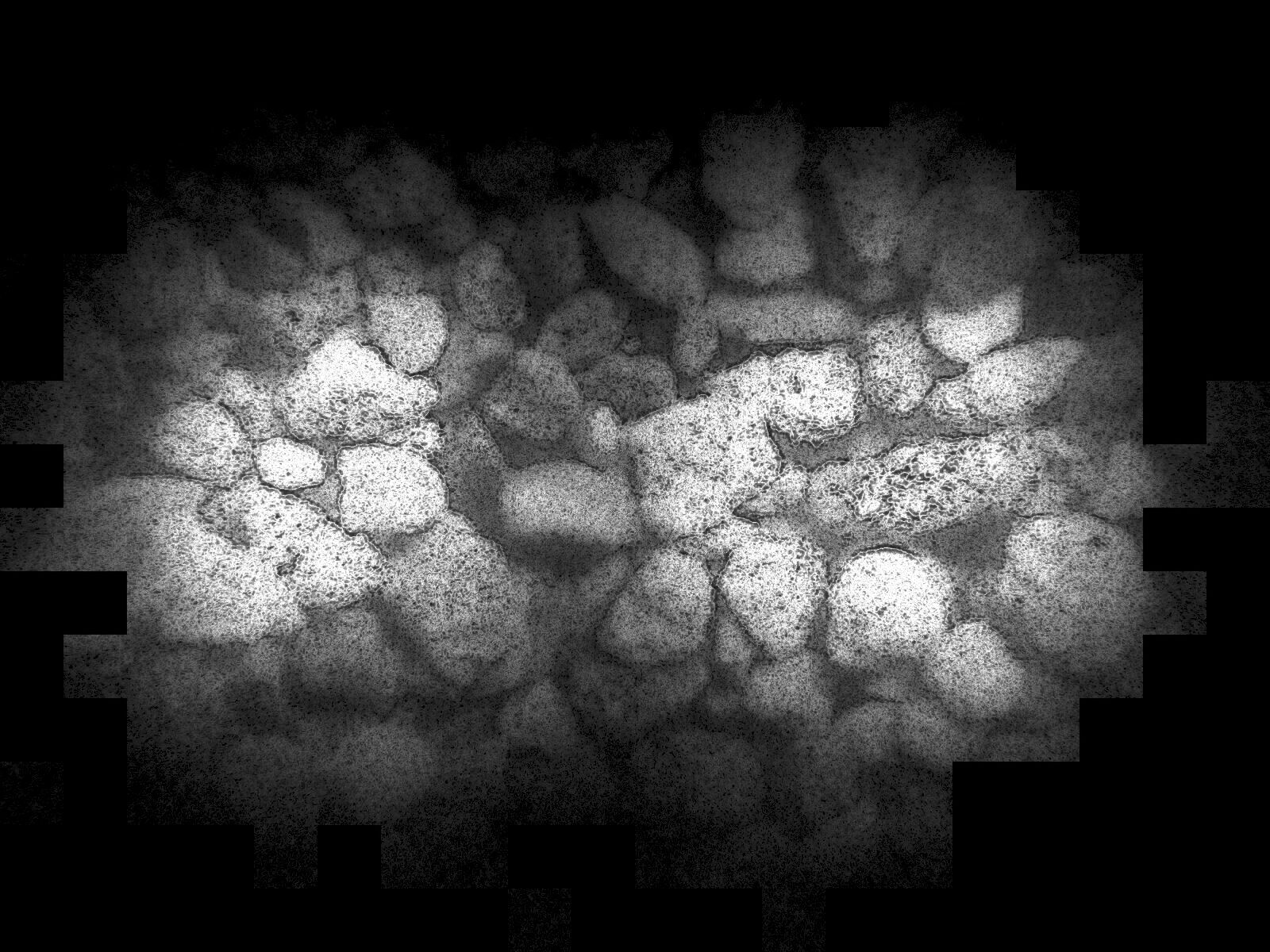}
	\includegraphics[width=0.3\linewidth]{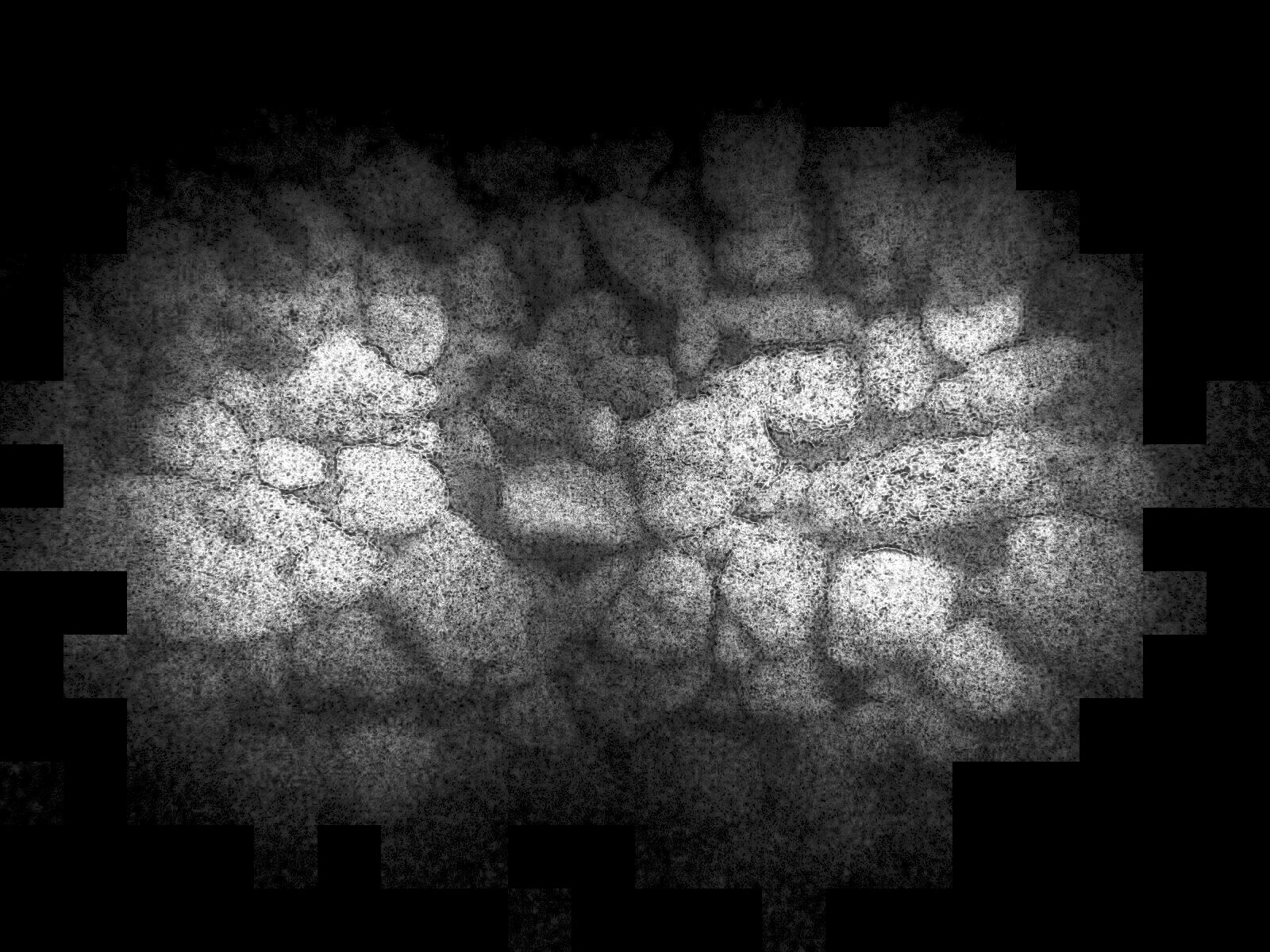} \\
	\includegraphics[width=0.3\linewidth]{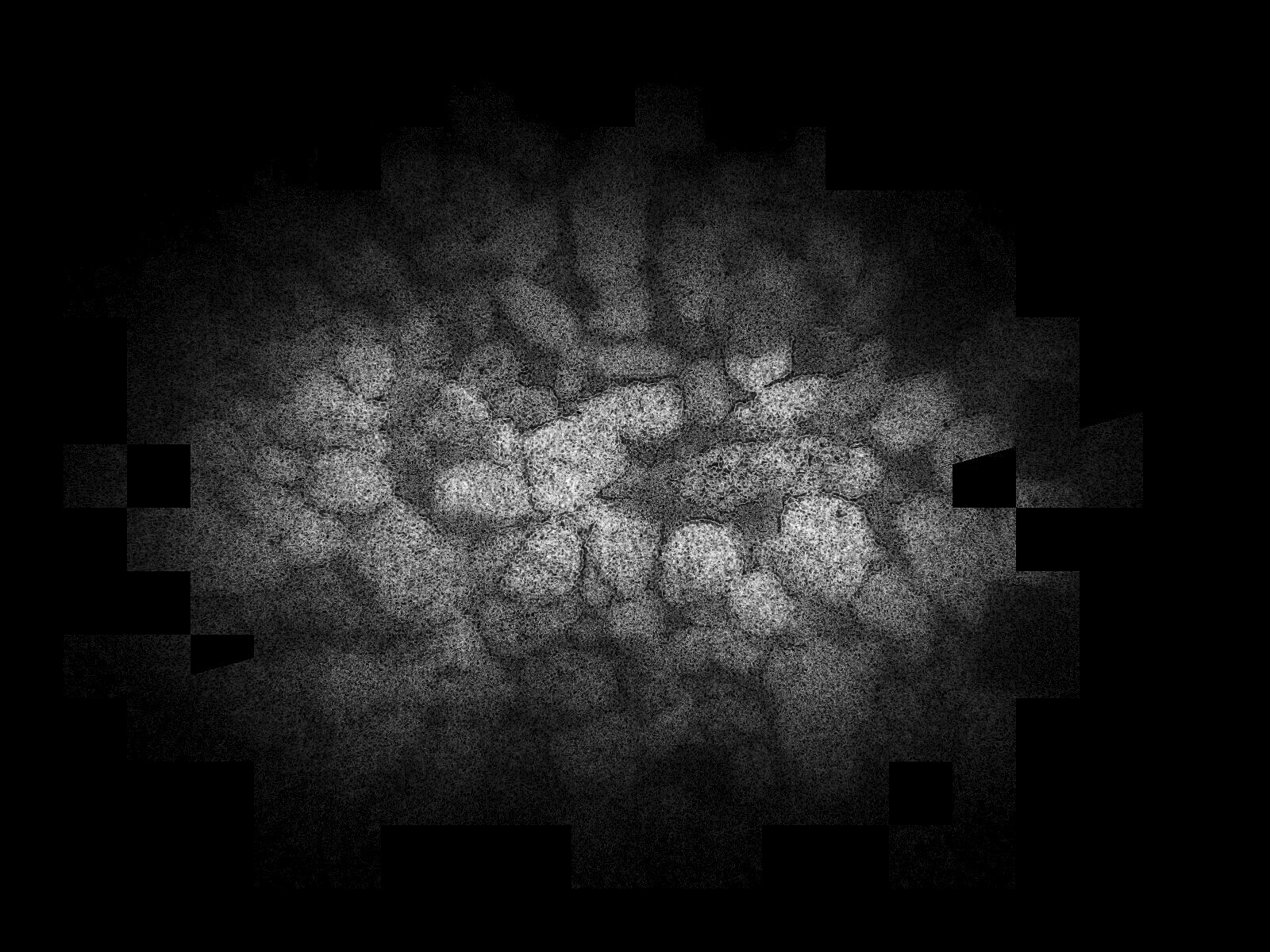}
	\includegraphics[width=0.3\linewidth]{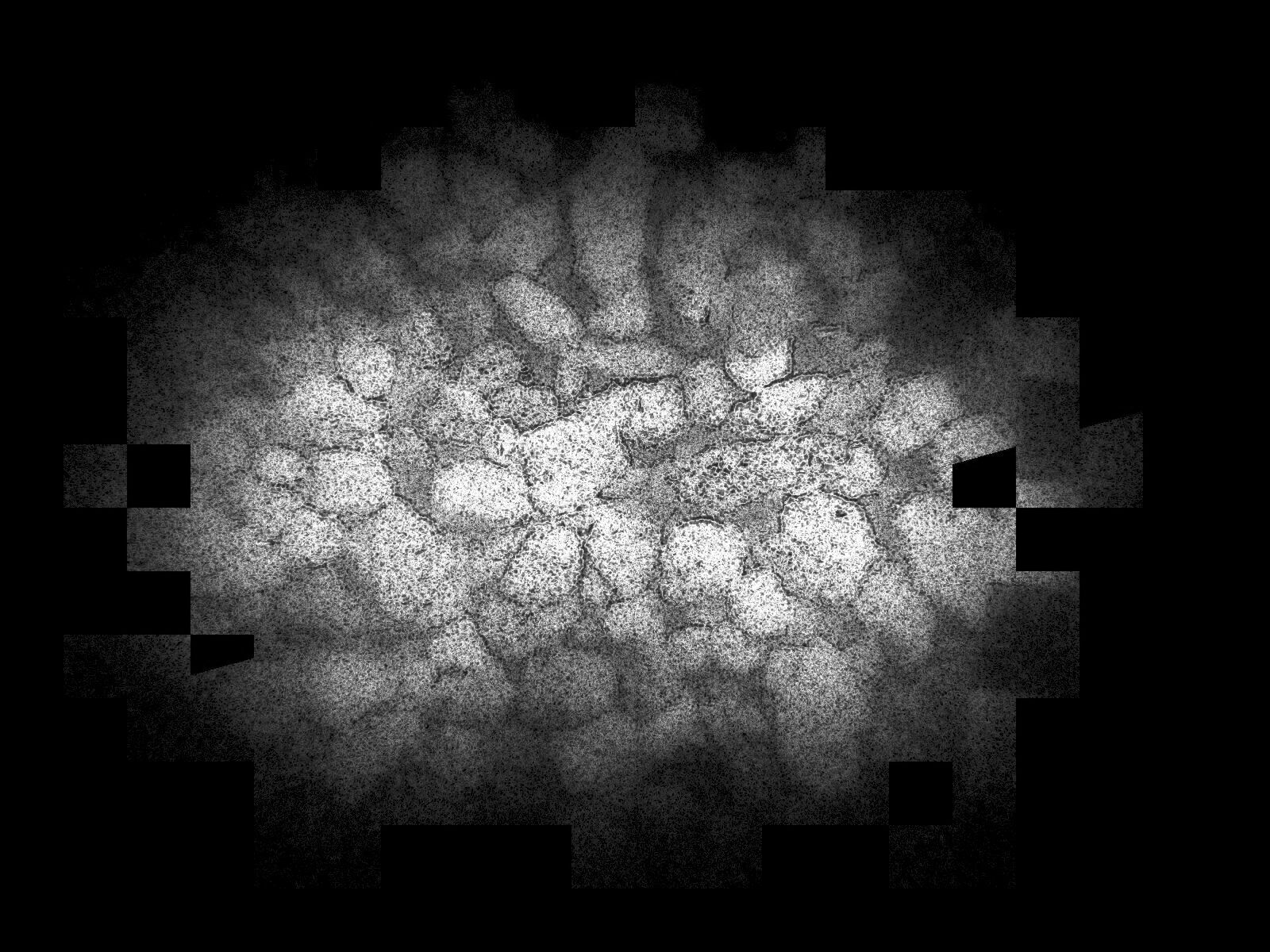}
	\includegraphics[width=0.3\linewidth]{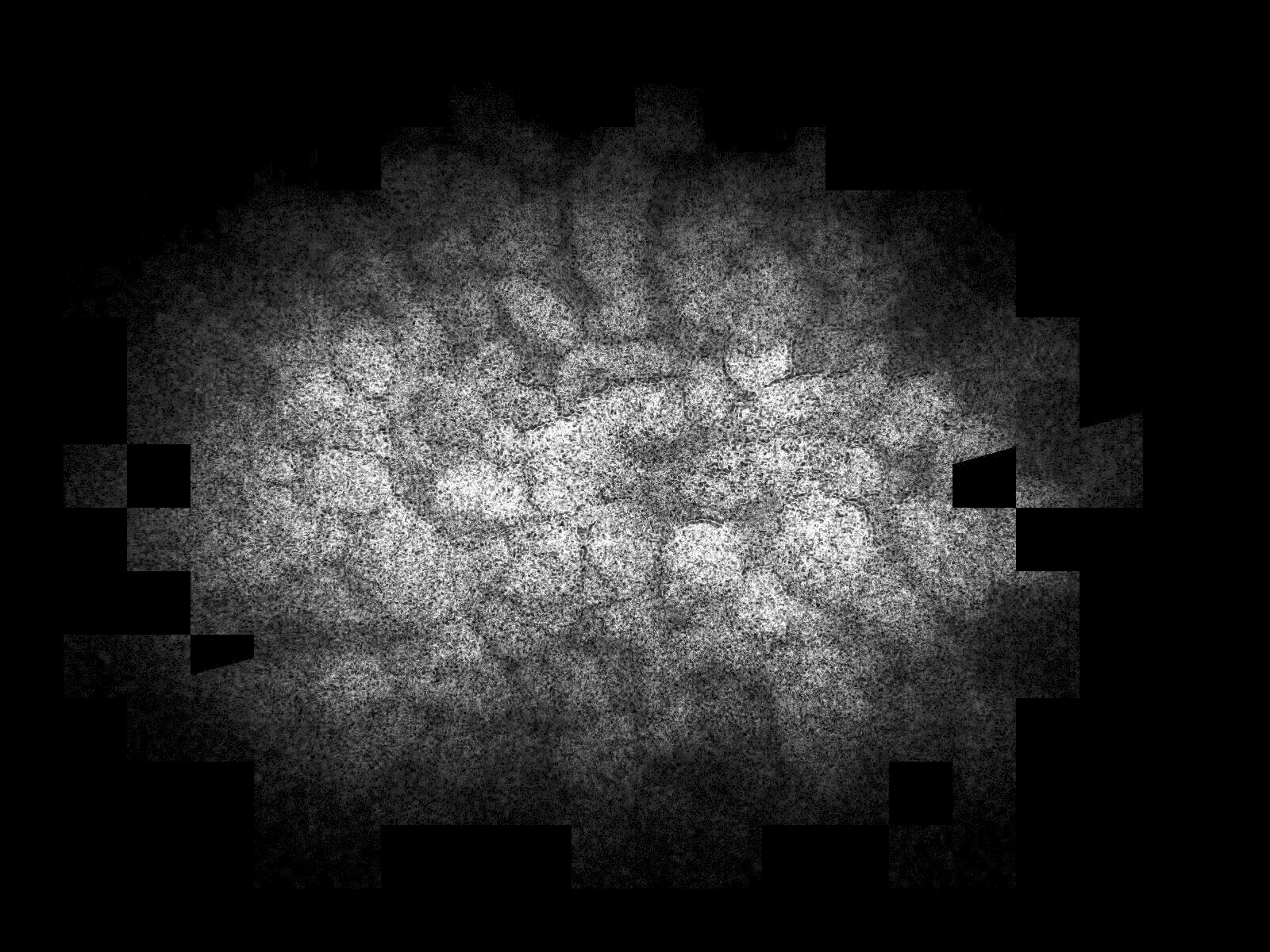}
	\caption{The confidence maps (R, G, B channels from left to right) for the corresponding restored images in Fig. 
		\ref{uw_calib_tank} showcase the level of confidence in the restoration process, with brighter values indicating higher reliability. These maps offer visual representations of the accuracy of restored colors at each pixel. Dark boundaries result from low illumination in those areas, leading to low SNR in both calibration and test images. Black patches in the confidence maps signify regions with insufficient information for parameter estimation or noisy color data, leading to a lack of confident estimation in the lookup table. Notably, the green and blue channels exhibit brighter values than the red channel due to the stronger absorption of red color by water, resulting in weaker signals and lower SNR in red channel. Additionally, some overexposed areas, mainly in the green and blue channels, display low confidence, while the red channel retains a higher level of confidence, as its intensities remain within an optimal range.}
	\label{uw_real_confidence}
\end{figure}

As shown if Fig. \ref{uw_calib_tank}, the estimated lookup table effectively describes the light patterns generated by the two artificial light sources. two light cones are widely separated at close distance and gradually merging to the center when distance increase. The separation and merging of the two light cones with distance are clearly visible, and a slight shift of the right-side light cone towards the image center, indicating a greater tilt of the right-side light source towards the camera (see Light 1 in Fig. \ref{uw_real_system}). These observations affirm the accurate estimation of the lookup table. The test images in the same figure showcase the successful removal of strong lighting patterns and underwater effects, resulting in the recovery of texture and consistent appearance. The presence of colorful boundaries in the restored images is attributed to insufficient information on the dark region in calibration images, leading to erroneous parameter estimation. Furthermore, the dark regions exhibit a noticeably low SNR, thereby exacerbating the noise in these areas. 
Corresponding confidence maps are also computed and displayed in Fig. \ref{uw_real_confidence}, provide a visual representation of the confidence level for each pixel in the restored images. Higher intensity values indicate stronger confidence in the accuracy of the restored colors for those pixels. The confidence value is influenced by both the original color information and the results of the lookup table estimation. Black patches in restored images indicate the absence of valid calibration data in those specific voxels, resulting in incorrect estimation of the lookup table parameters. Additionally, certain pixels may be overexposed (mostly in blue and green channels), such as the bright spot in the first test image, causing low confidence values in the blue and green channels, while higher confidence is still maintained in the red channel for these pixels. 

\begin{figure}
	\centering
	\includegraphics[width=0.32\linewidth]{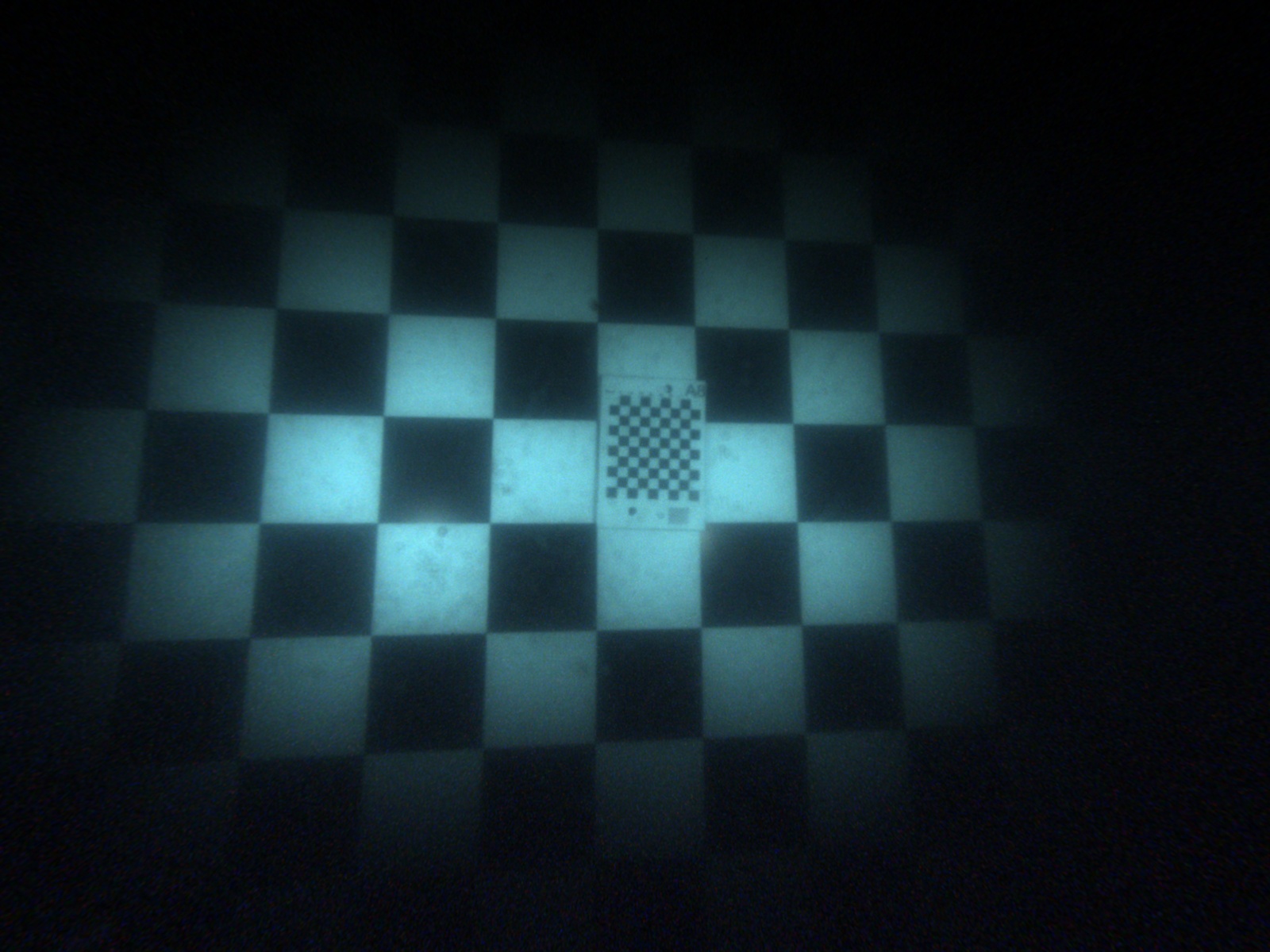}
	\includegraphics[width=0.32\linewidth]{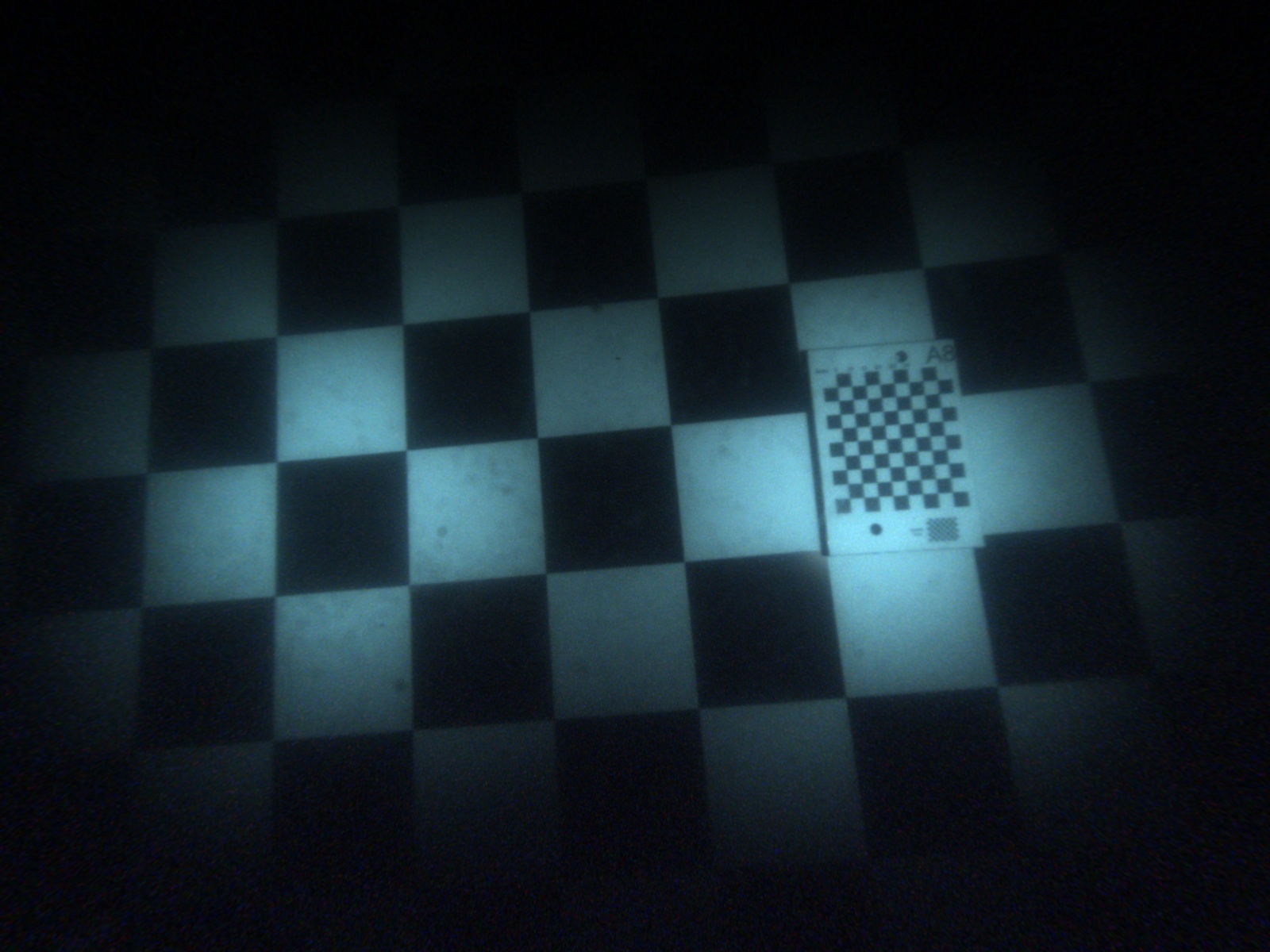}
	\includegraphics[width=0.32\linewidth]{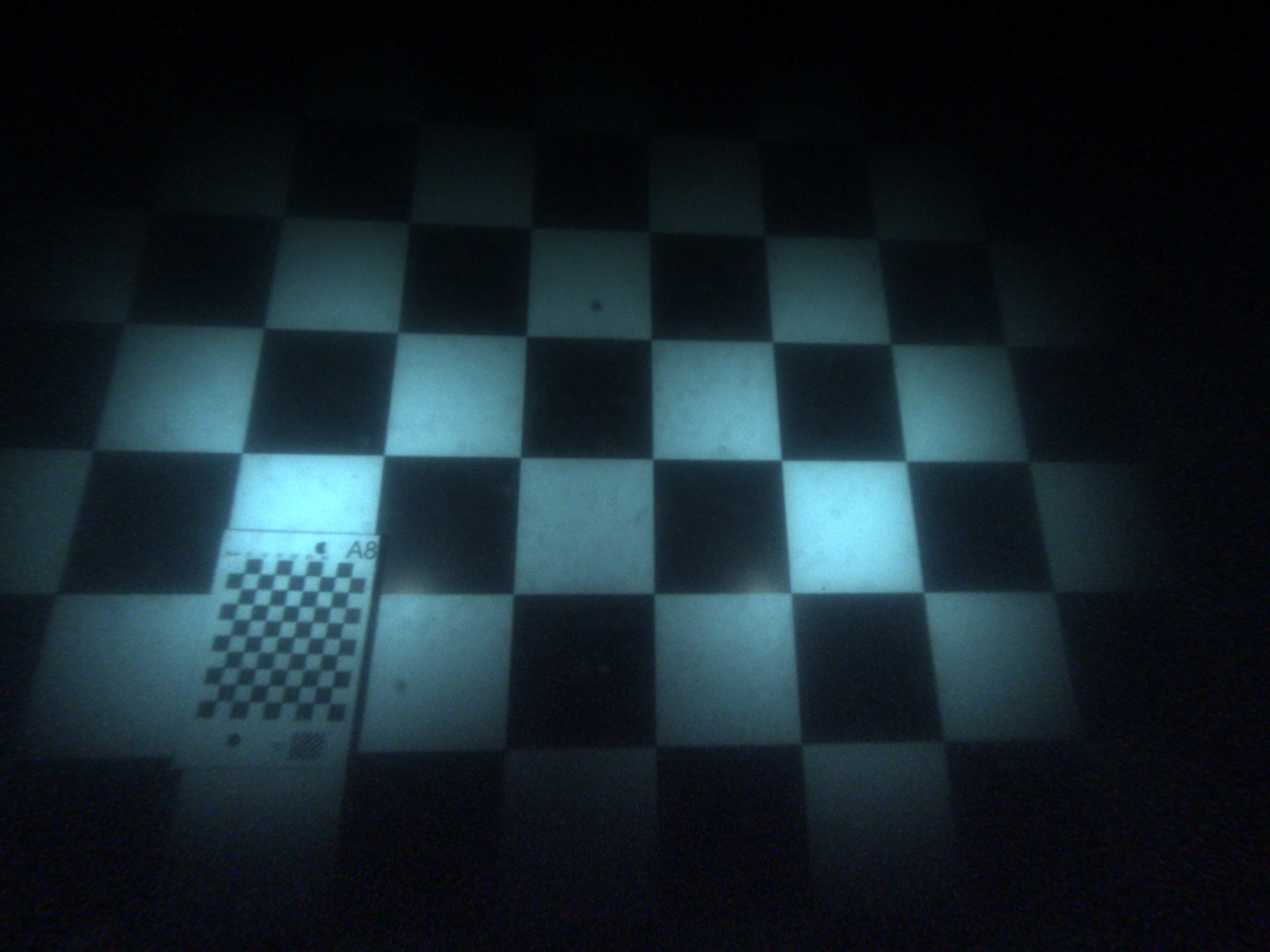}\\
	\includegraphics[width=0.32\linewidth]{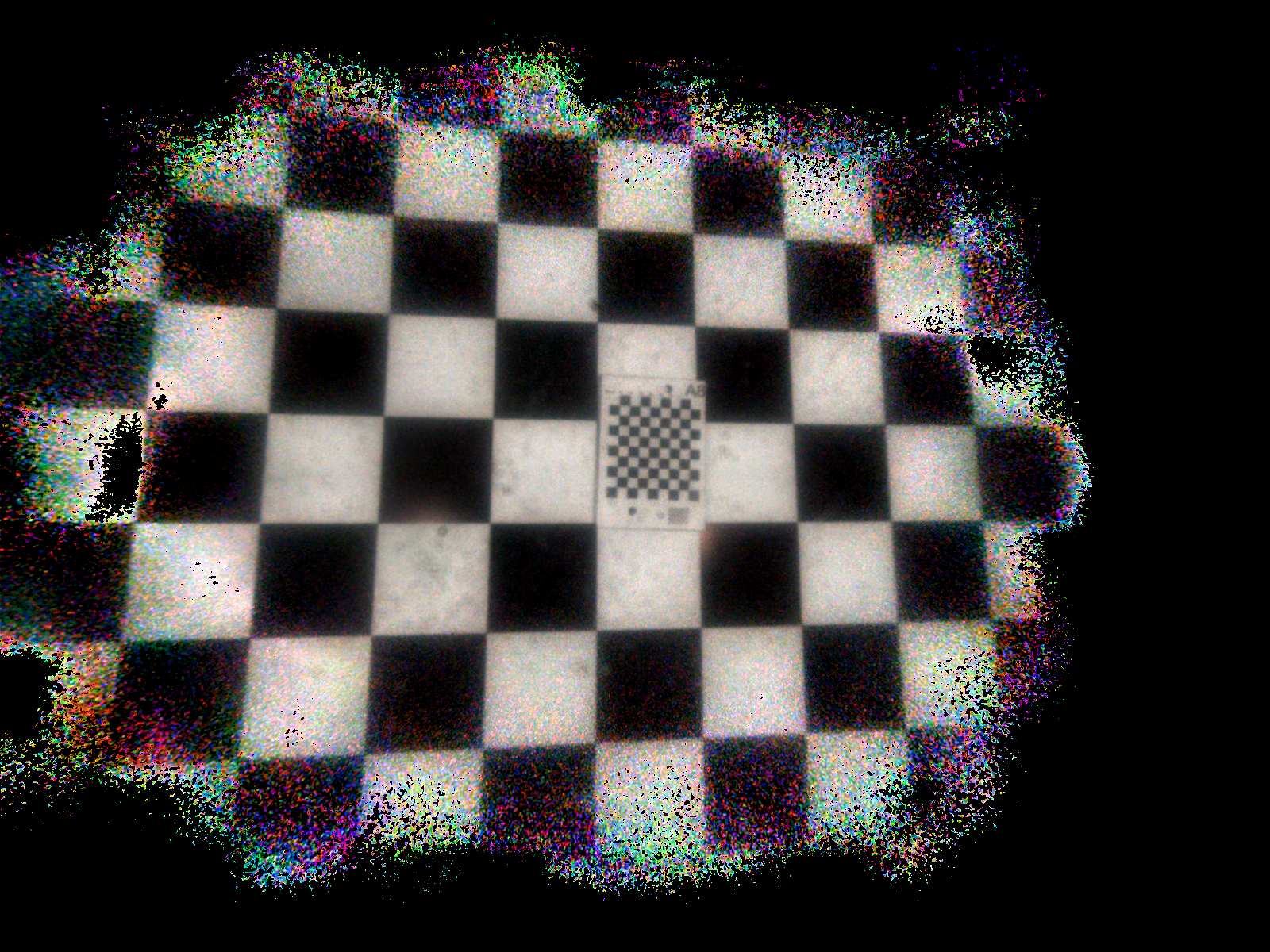}
	\includegraphics[width=0.32\linewidth]{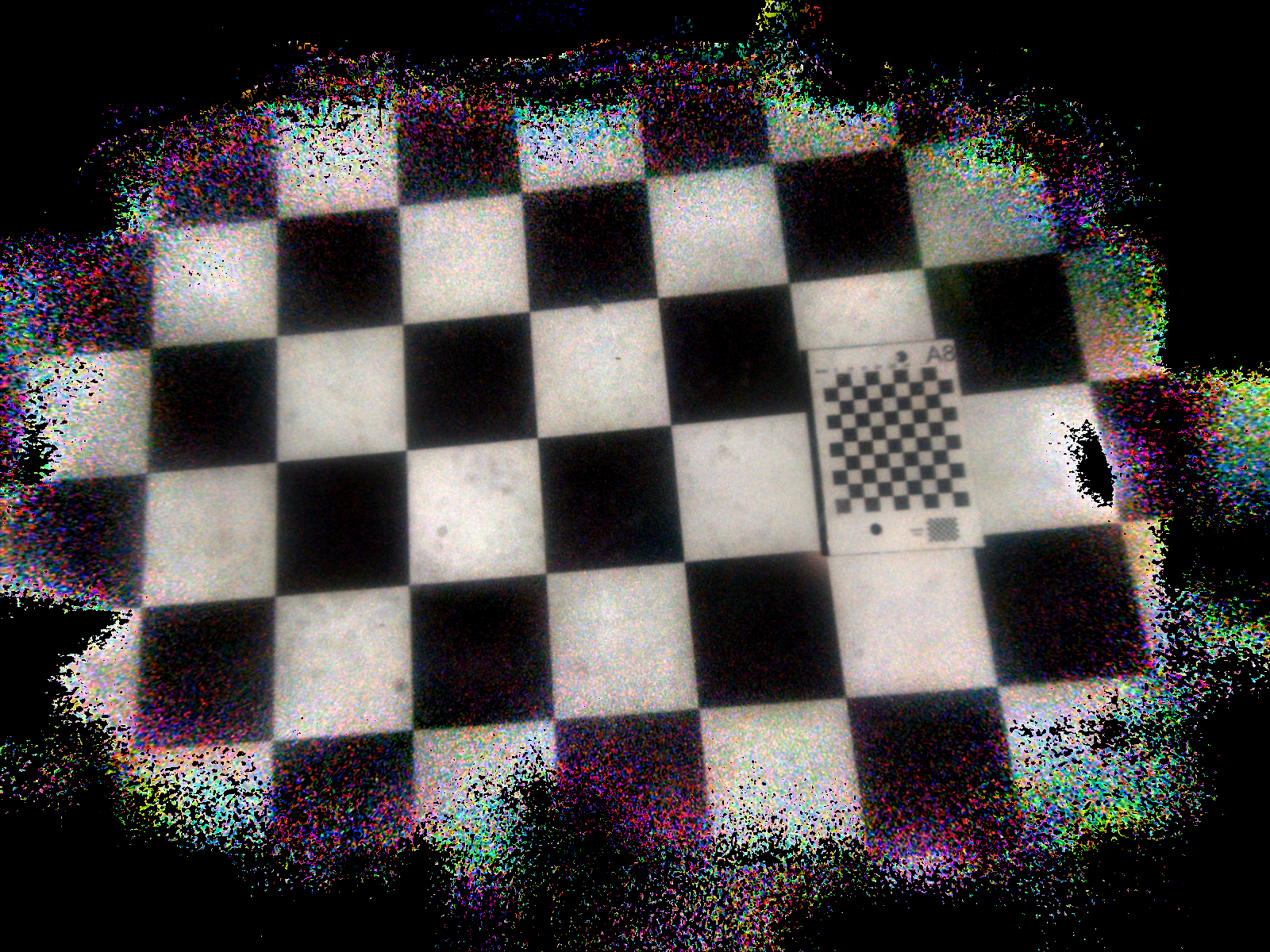}
	\includegraphics[width=0.32\linewidth]{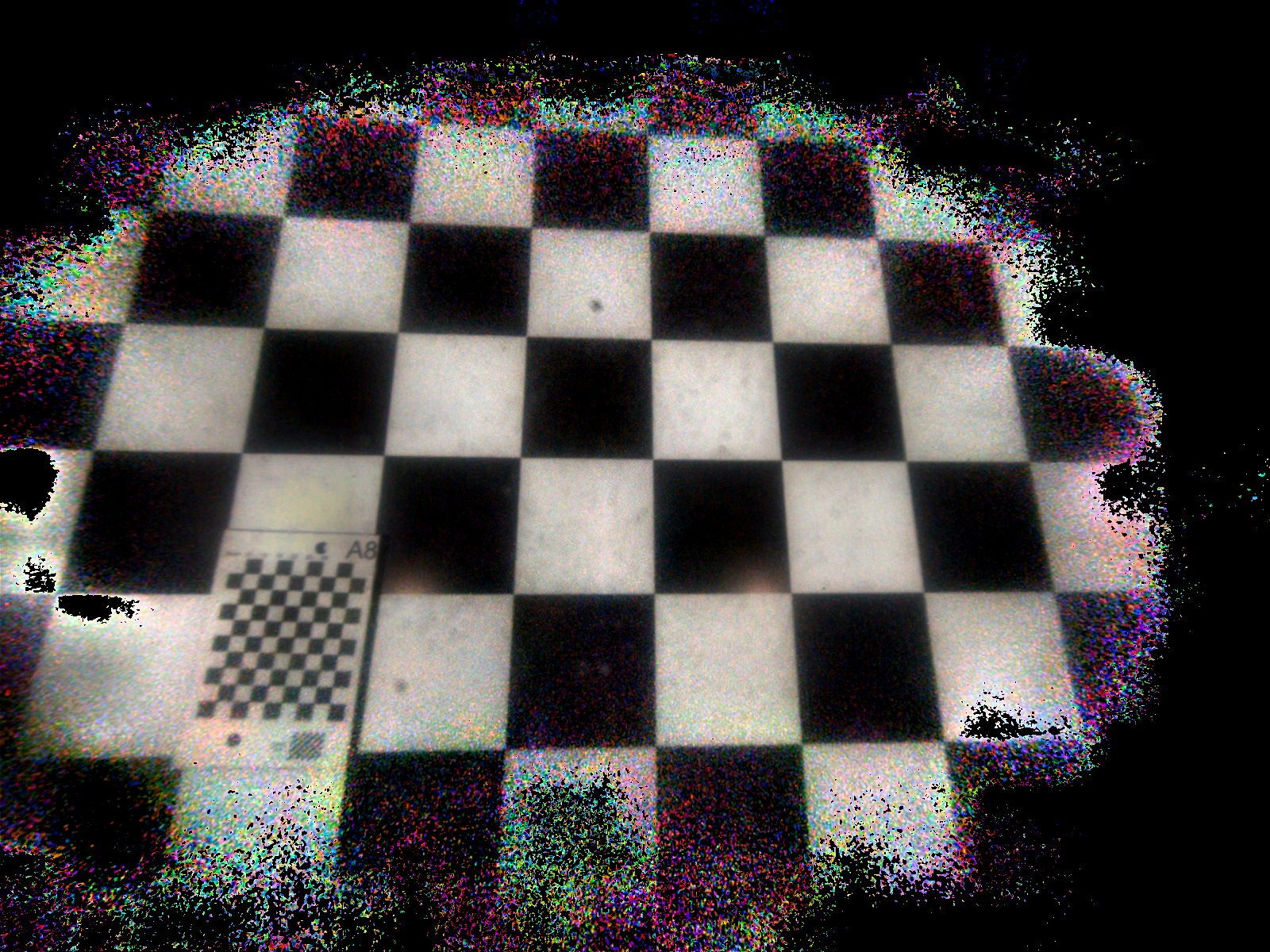}
	\caption{Example chessboard images and their restoration results. The top row displays the original underwater images, while the bottom row shows the corresponding restored images with true color, where artificial lighting patterns and water effects have been successfully removed. }
	\label{uw_real_chessboard_results}
\end{figure}

The effectiveness of our approach is further demonstrated in Fig. \ref{uw_real_chessboard_results}, where the restored chessboard images clearly exhibit the removal of complex dynamic light patterns and the recovery of image abedo. 

\subsection{Parameter Estimation from Correspondences}
%In real ocean applications, capturing calibration data at each voxel position inside the viewing frustum using a calibration board is challenging, especially with wide field of view cameras. This section introduces the method for estimating lookup table parameters primarily based on image correspondences. 

%In this case, constraints are mostly from correspondences, it has the risk to reach zero during optimization, additional weighting factor is needed to decrease the impact of correspondences constraints in the whole optimization system. The factor is depend on the number of correspondences constraints compare to the number of other types of constraints.

%This section introduces another strategy for estimating lookup table parameters primarily based on image correspondences. 

Previous experiments have demonstrated the viability of estimating the lookup table for underwater image restoration when utilizing known color calibration objects. In such instances, known color constraints serve as the primary source of information for estimating the lookup table parameters, with other constraints offering supplementary information in regions not covered by the known color constraints. 
In this section, we delve into the scenario where known color calibration objects are unavailable and explore the potential of leveraging correspondence information from multi-view images of arbitrary scenes to calibrate the lookup table. In this case, the constraints mainly arise from correspondences. 

\begin{figure}
	\centering
	\includegraphics[width=1.0\linewidth]{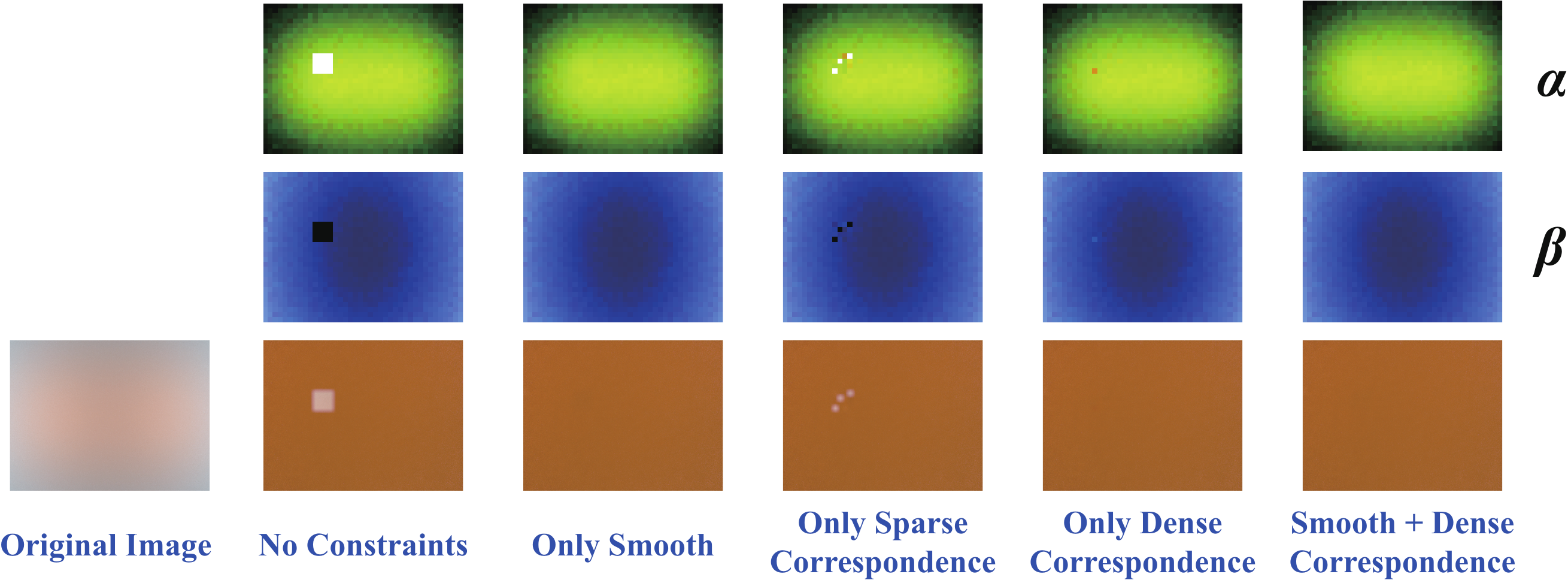}
	\caption{Comparison of lookup table estimation results using different supplementary constraints.  From left to right: (1)  Original turbid underwater test image. (2) Lookup table estimation results with only known color constraints, a deliberately chosen 4$\times$4 test region where all constraints has been removed. (3) Results with only smooth constraints in the test region. (4) Results with only sparse correspondence constraints within the test region, the correspondence constraints linking the unconstrained voxels inside the region with the constrained voxels outside. (5) Results with only dense correspondence constraints within the test region. (6) Result with both smooth and dense correspondence constraints integrated into the lookup table parameters estimation. } 
	\label{uw_correspondence_impact}
\end{figure}

Before delving into the methodology of the correspondence-based approach, it is essential to revisit the role of correspondence constraints within the known color-based approach and assess their influence on the estimation of lookup table parameters. The simulated turbid water dataset used previously is employed here to demonstrate the impact of correspondence constraints. In order to showcase this impact, we focus on a specific 4$\times$4 region within one of the slabs of the lookup table. Within this region, all known color information was intentionally remove. If we were to attempt the direct estimation of the lookup table without supplementary constraints, the parameters within this region would remain unaltered throughout the optimization process (refer to Fig. \ref{uw_correspondence_impact} second column). 
When solely employing smoothness constraints as supplementary factors, the empty region would be interpolated using information from neighboring regions with known color constraints (as seen in the third column of Fig. \ref{uw_correspondence_impact}). The calibrated values would gradually spread to the uncalibrated region over successive iterations. For a 4$\times$4 area, this coverage would occur within just two iterations. 
On the other hand, when using only correspondence constraints, calibrated values from outside regions which are constrained by known colors would integrate with the uncalibrated parameters within the test region to form each correspondence constraint. In the uncalibrated area, the super-pixel centers are extracted and utilized to establish the correspondence constraints. Only those centers that have correspondences outside the test region with known parameters would be constrained with a unique solution, while other voxels within this region would possess unconstrained estimated values. When denser super-pixels are extracted within the test region, a greater number of correspondences are generated, resulting in more voxels' parameters being estimated with unique solutions. The influence of these correspondence constraints and their effects on parameter estimation are depicted in the fourth and fifth columns of Fig. \ref{uw_correspondence_impact}. 
The last column displays the outcomes obtained by integrating smooth and correspondence constraints within the test region. Unlike the outcomes solely based on smooth constraints, which involve straightforward value interpolation from neighboring voxels, and those relying solely on correspondence constraints, which may leave uncovered voxels, the integrated approach offers a more comprehensive and precise estimation of the lookup table within the test region. 

With the known color constraints, it's noteworthy that half of the unknown parameters in each correspondence constraint are already resolved. This simplifies the process of achieving a unique solution for the equation system, given that half of the unknown parameters are already estimated. 
However, when exclusively solving the equation system relying on correspondence constraints, two distinct general solutions can be identified in Eqt. \ref{eqCorrespondence_int}. 
The first solution is $\alpha_{1,2} = 0$. This implies that when filming an object without any illumination, the correspondence constraints are automatically satisfied. The second solution arises when $I_1 = \beta_1$ and $I_2 = \beta_2$, which signifies the filming of a black body object and the correspondence constraints are again fulfilled. 
To prevent all $\alpha$ values from becoming zero, an additional normalization constraint was imposed on them ($\sum^{n}_{i=0}{\alpha_{i}} = 1.$).
Similarly, in order to avoid $\beta_{i}$ from becoming the observed color, it is necessary for each voxel to capture multiple distinct colors during the data acquisition. Moreover, considering the potential errors in the color observations, if each voxel captures similar colors, the ambiguities still remains in the equation system. To mitigate this, it is crucial to capture images in complex scenes with a diverse range of colors. This ensures that each voxel obtains sufficient color observations, enabling the accurate estimation of lookup table parameters. 
After estimating the lookup table parameters, all $\alpha$ values are still normalized, requiring them to be scaled to the appropriate scale. The scaling factor can be directly estimated from a single voxel with an absolute $\alpha$ value, which is obtained from known color constraints.   

\subsection{In-air Calibration Predominantly Based on Correspondences}
To validate the correspondence-based lookup table parameter estimation approach, we conducted a test in a simulated in-air scenario. As mentioned above, achieving a unique solution for the correspondence constraints requires diverse color observations.  Therefore, a 3D plane with random unknown color patches texture was used as the object, providing a wide range of colors to satisfy the correspondence constraints. 
To simplify the experimental setup, we ensured that all observed points were within one slab by simulating images from a fixed distance and viewing direction to the textured plane. The camera was constrained to shift and rotate on a virtual plane above the object, while a co-moving point light source was placed in front of the camera. 

\begin{figure}
	\centering
	\includegraphics[width=1.0\linewidth]{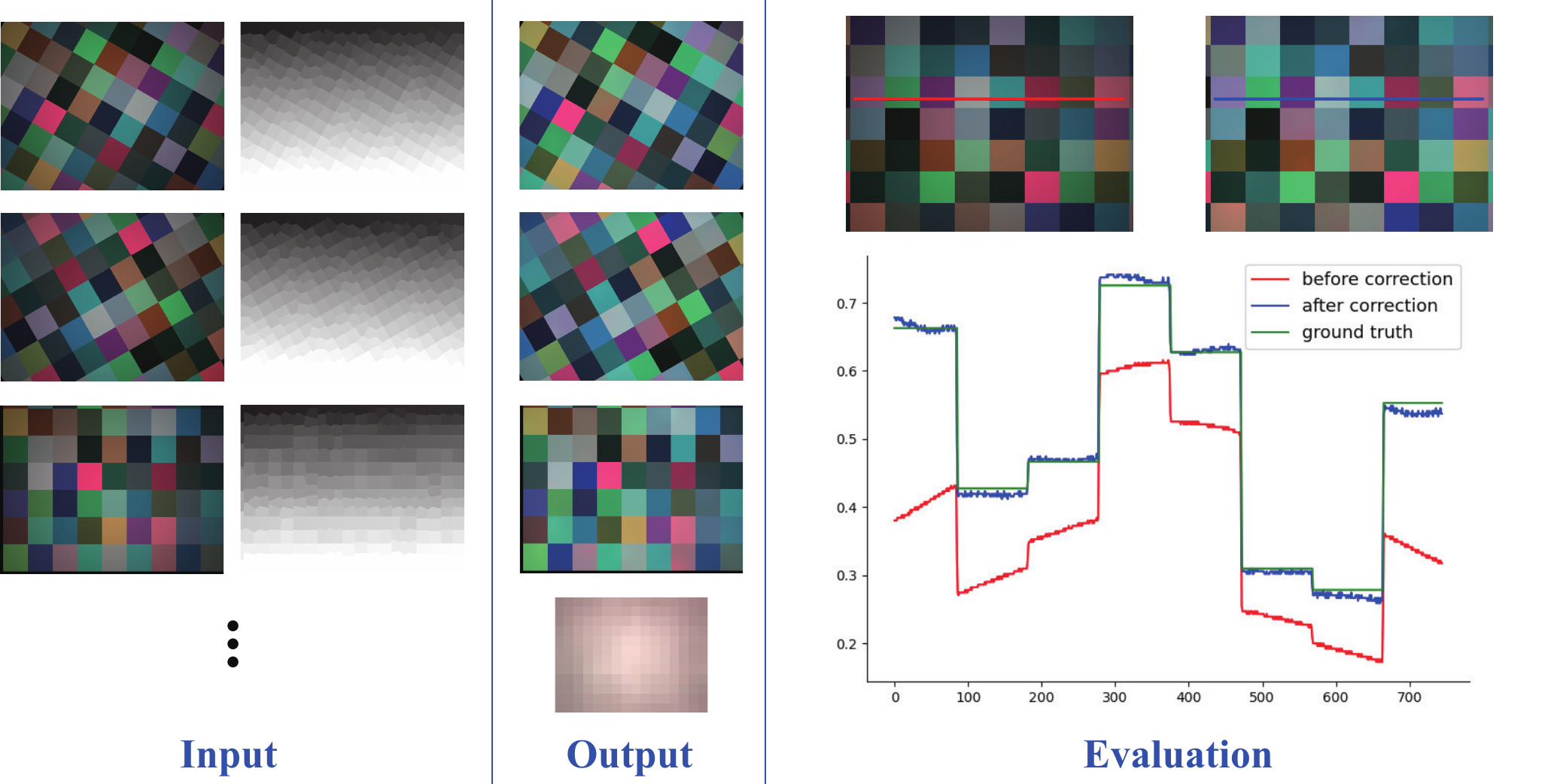}
	\caption{Experiment results on the simulated in-air dataset primarily utilizing image correspondences for image restoration. 
		Left: Test images of a colorful plane used in the image restoration experiment, with 300 super pixels extracted from each image to construct correspondence constraints. These images exhibit uneven illumination due to a co-moving point light source. 
		Middle: Corresponding restored images obtained using the estimated one slab lookup table (shown in bottom) from the correspondence constraints. 
		Right: Evaluation of the restoration result. The top two images show examples before and after restoration, while the bottom figure displays the blue channel intensities sampled along the lines in these images. In the original images (in red), noticeable gradients are observed in each patch due to point light shading, and the values significantly deviate from the ground truth intensities (in green).  After the correction, the intensities (in blue) become relatively constant in each patch, closely matching the ground truth values. This demonstrates the successful removal of uneven illumination and the accurate restoration of color in the images.
	}
		
	\label{in_air_correspondence}
\end{figure}

Eighteen test images were generated to estimate the one slab lookup table parameters. 
As backscatter ($\beta$) is negligible in in-air images, the focus was solely on estimating the transmission factor $\alpha$ for each voxel.
Fig. \ref{in_air_correspondence} illustrates the entire restoration procedure: 300 super pixels were extracted from each input image, and with the known extrinsics of each image, the center of each super pixel was projected into the corresponding paired image to construct the correspondence constraints. 
Based on these constraints, the one slab lookup table (size: 16×12×1) with normalized $\alpha$ values was estimated. Subsequently, a single point from one of the images was selected, and its true color served as the scale factor to compute the absolute value for the corresponding voxel. The entire $\alpha$ values in the lookup table were then re-scaled by this voxel. Using the re-scaled lookup table, the colors of all input images were corrected. The resulting corrected images demonstrated the successful removal of uneven illumination. 
Furthermore, the plotted intensity distributions along the lines in the images, before and after the correction, indicated relatively constant intensity in each patch of the corrected images. The quality of the estimated lookup table parameters for each voxel depended on the observed intensities, with higher robustness achieved when there were more observed colors and greater diversity among these colors. 

In the underwater scenario, theoretically, it's possible to attain a unique solution for lookup table estimation when an ample number of correspondences are provided within the same voxel. However, each correspondence constraint encompasses four unknown parameters intertwined through multiplication. To achieve sufficient constraints for every voxel, an extraordinarily dense observation and an exceedingly complex scene with diverse colors are required. Especially when observations are prone to errors, we encountered a challenge that the optimizer is difficult to distinguish whether the effects stem from the $\alpha$ or $\beta$ terms. This predicament remains an unresolved question that warrants further investigation.

%-------------------------------------------------------------------------
\section{Conclusion}

This paper proposes a general underwater image formation model and presents a novel and versatile solution for underwater image restoration based on a 3D lookup table. This approach overcomes the drawbacks of traditional methods based on classical underwater image formation models and effectively handles the challenges posed by complex water and lighting effects. 
Extensive experiments on simulated and real-world datasets validate the effectiveness of our approach. The results demonstrate its ability to restore the true albedo of objects while mitigating the influence of lighting and medium effects. This capability is particularly valuable for underwater large scale 3D reconstruction and mapping tasks, where accurate and consistent color information is essential. Moreover, we have shown that our method can be readily extended to other scenarios, including in-air cases with artificial illumination.

%-------------------------------------------------------------------------
\section*{Acknowledgements}
This publication has been funded by the German Research Foundation (Deutsche Forschungsgemeinschaft, DFG) Projektnummer 396311425, through the Emmy Noether Programme, and the European Union’s
Horizon 2020 research and innovation programme under Grant
Agreement No 101000858 (TechOceanS). 

%-------------------------------------------------------------------------
\bibliography{references.bib}

\end{document}